\documentclass[10pt,journal,compsoc]{IEEEtran}

\ifCLASSOPTIONcompsoc
  \usepackage[nocompress]{cite}
\else
  \usepackage{cite}
\fi
\usepackage{times}
\usepackage{epsfig}
\usepackage{graphicx}
\usepackage{amsthm}
\usepackage{amsmath}
\usepackage{amssymb}
\usepackage{subfigure}
\usepackage{epstopdf}
\usepackage{algorithm}
\usepackage{algorithmic}
\usepackage{multirow}
\usepackage{color}
\usepackage{ragged2e}
\usepackage{subfig}
\usepackage{caption}
\usepackage{footnote}
\usepackage{etoolbox}

\DeclareMathOperator*{\argmax}{argmax}

\theoremstyle{plain}

\ifCLASSINFOpdf
\else
\fi
\begin{document}
%
\title{Object Discovery via Cohesion Measurement}
%
%
%
%

\author{Guanjun~Guo,
        Hanzi~Wang*,~\IEEEmembership{Senior Member,~IEEE,}
        Wan-Lei~Zhao,~\IEEEmembership{} \\
        Yan~Yan,~\IEEEmembership{Member,~IEEE,}
        Xuelong~Li,~\IEEEmembership{Fellow,~IEEE}
\IEEEcompsocitemizethanks{
\IEEEcompsocthanksitem G.~Guo, H.~Wang, W.~Zhao and Y.~Yan are with the Fujian Key Laboratory of Sensing and Computing for Smart City, and the School of Information Science and Engineering, Xiamen University, Xiamen, 361005, Fujian, P. R. China.\protect
 ~(E-mail: gjguo@stu.xmu.edu.cn; hanzi.wang@xmu.edu.cn; wlzhao@xmu.edu.cn; yanyan@xmu.edu.cn).

\IEEEcompsocthanksitem X. Li is with the Center for Optical Imagery Analysis and Learning, State Key Laboratory of Transient Optics and Photonics, Xi'an Institute of Optics and Precision Mechanics, Chinese Academy of Sciences, Xi'an, 710119, Shaanxi, P. R. China.\protect
 ~(E-mail: xuelong$\_$li@opt.ac.cn).
 \IEEEcompsocthanksitem *~ The corresponding author.\protect}
\thanks{}}

%
%

\markboth{}%
{Shell \MakeLowercase{\textit{et al.}}: Bare Advanced Demo of IEEEtran.cls for Journals}
%



\IEEEtitleabstractindextext{%
\noindent\begin{justify}
\begin{abstract}
Color and intensity are two important components in an image. Usually, groups of image pixels, which are similar in color or intensity, are an informative representation for an object. They are therefore particularly suitable for computer vision tasks, such as saliency detection and object proposal generation. However, image pixels, which share a similar real-world color, may be quite different since colors are often distorted by intensity. In this paper, we reinvestigate the affinity matrices originally used in image segmentation methods based on spectral clustering. A new affinity matrix, which is robust to color distortions, is formulated for object discovery. Moreover, a Cohesion Measurement (CM) for object regions is also derived based on the formulated affinity matrix. Based on the new Cohesion Measurement, a novel object discovery method is proposed to discover objects latent in an image by utilizing the eigenvectors of the affinity matrix. Then we apply the proposed method to both saliency detection and object proposal generation. Experimental results on several evaluation benchmarks demonstrate that the proposed CM based method has achieved promising performance for these two tasks.
\end{abstract}
\end{justify}
}

\maketitle

\IEEEdisplaynontitleabstractindextext

%
\IEEEpeerreviewmaketitle

\ifCLASSOPTIONcompsoc
\IEEEraisesectionheading{\section{Introduction}\label{sec:introduction}}
\else
\section{Introduction}
\label{sec:introduction}
\fi

%
%
%
%
\IEEEPARstart{O}{bject} detection is one of fundamental tasks in computer vision. In the literature, this problem is conventionally addressed by applying a classifier on all possible hypotheses (i.e., all object candidate regions in an image). At the beginning stage, the candidate regions are exhaustively generated to cover all possible locations and scales based on the sliding window strategy (e.g.,~\cite{Viola2001,HOG2005,YangC14}). However, the sliding window strategy is computationally expensive. In order to reduce the number of the candidate regions, image segmentation based methods are adopted, which generate much less number of class-independent object proposal windows~\cite{SelectiveSearch13,xuelong2017}. As a consequence, a classifier (i.e., an object detector) only focuses on the proposal windows to identify the objects in an image. This kind of methods includes R-CNN~\cite{Rcnn2014}, SPP-net~\cite{Sppnet2014}, etc.
With the support of proposal windows based on image segmentation, the task of object detection has been relieved from the heavy burden of exhaustive detection, whereas the performance of those methods becomes sensitive to the accuracy of image segmentation. Unfortunately, perfect image segmentation is hard to achieve in general.

\begin{figure}
\begin{center}
   \subfigure[]{
   \begin{minipage}[b]{0.15\textwidth}
   \includegraphics[width=1\textwidth]{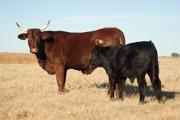}\\
   \includegraphics[width=1\textwidth]{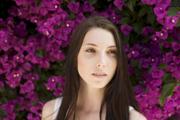}\\
   \includegraphics[width=1\textwidth]{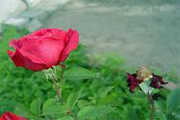}
   \end{minipage}}
   \subfigure[]{
   \begin{minipage}[b]{0.15\textwidth}
   \includegraphics[width=1\textwidth]{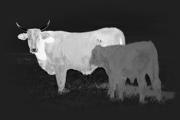}\\
   \includegraphics[width=1\textwidth]{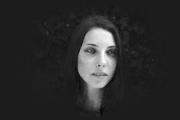}\\
   \includegraphics[width=1\textwidth]{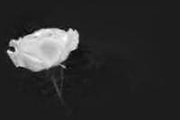}
   \end{minipage}}
    \subfigure[]{
   \begin{minipage}[b]{0.15\textwidth}
   \includegraphics[width=1\textwidth]{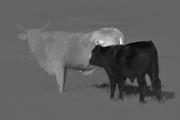}\\
   \includegraphics[width=1\textwidth]{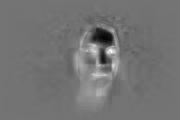}\\
   \includegraphics[width=1\textwidth]{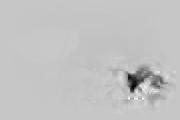}
   \end{minipage}}
\end{center}
   \vspace{-0.451cm}
   \caption{\label{overviewFig}The proposed object discovery method is illustrated to find objects in an image. (a) The input images. (b)-(c) The detected objects corresponding to the first two eigenvectors ranked by their eigenvalues in descending order, where the eigenvalues are the cohesion measurement values.}
\end{figure}
Aiming to circumvent the ill-posed problem of image segmentation, several other generic object proposal generation methods are proposed. For example, the objectness methods~\cite{Objectness2012,BingObj2014} are able to generate category-independent proposal windows efficiently, and they can achieve high recall. However, the major disadvantage of these methods is that they are unable to generate a bounding box for an object as precisely as a segmentation based method. Compared with the methods based on image segmentation and object proposal, saliency detection~\cite{FangTIP12,fang2014,cao2014} makes a better trade-off between the computational cost and the accuracy for estimating the bounding boxes of objects. The principle of saliency detection is that it only selects the regions which draw the attention of human beings at their first sight~\cite{VisualAttention2013,huazhu2013}. Similar to object detection methods, saliency detection has been adopted as a pre-processing step in various applications. However, most saliency detection methods only extract salient foreground objects from an image, which hinders themselves from being adopted to detect all the objects in an image. Moreover, finding all the objects in an image is beyond the scope of saliency detection.

In this paper, we reinvestigate spectral clustering for image segmentation, and formulate a novel and effective affinity matrix for object discovery. Then, we apply the eigenvectors of the affinity matrix to the problems of saliency detection\footnote{The proposed method is applied to the task of salient object detection in this paper. For simplicity, we use the term of saliency detection.} and object proposal generation. The intuition behind the proposed method is based on a common observation that an object region usually distinguishes itself from its surroundings mainly due to color and intensity differences. In real-world images, image pixels on the same object region share similar real-world color values. Meanwhile, due to variations in lighting conditions and shadow effects, image pixels of an object region in an image are usually not concentrated in the RGB color space. Instead, they are usually distributed within an elongated ellipse region in the RGB space. Different object regions (with different dominant colors) in an image form different elongated ellipse regions with different orientations. Omer and Werman~\cite{ColorLine2004} are the first who observed this case. In their work, the elongated ellipse regions are called as color lines, and this phenomenon is called as the color line phenomenon. The color lines are essentially caused by color distortions inside different object regions. Because of the color lines phenomenon, the image pixels, which share a dominant color, may be quite far away from each other when being measured with the Euclidean distance metric. As a result, a heuristical way to segment different latent object regions in an image is to propose an image segmentation method based on color lines. Then objects contaminated by color distortions can be segmented conveniently based on the obtained object regions. On the other hand, image segmentation methods based on spectral clustering (such as~\cite{Shi2000,Freeman1998}) also face the problem of color distortions. However, the affinity matrices used in those spectral clustering based methods do not consider the influence of color distortions. That is the reason why some image segmentation methods based on spectral clustering do not work well on natural scene images.

To address the above issues, a novel affinity matrix, which is robust to color distortions, is proposed in this study. In addition, a cohesion measurement for object regions is also adopted in the proposed affinity matrix. Generally speaking, different object regions obtain different cohesion measurement values. Thus, the problem of object discovery can be treated as the problem of detecting different local maxima of the cohesion measurement for an image. Different from traditional spectral clustering methods (e.g.,~\cite{Freeman1998,Shi2000,Ng01}), the eigenvalues of the proposed affinity matrix can be used to measure the cohesion of object regions, and the corresponding eigenvectors contain important information of the objects. Fig.~\ref{overviewFig} outlines the general idea of the proposed method. The original images are shown in Fig.~\ref{overviewFig}(a). The detected salient objects, which correspond to the top-2 ranked eigenvectors of the affinity matrix, are shown in Fig.~\ref{overviewFig}(b)-(c), respectively.

As the major contribution of this paper, a novel method, based on a new Cohesion Measurement (CM), is proposed to discover objects latent in an image. Inspired by the concept of color lines, some existing image segmentation methods based on spectral clustering are reinvestigated and a novel affinity matrix, which is robust to color distortions, is proposed in this study. Since different eigenvectors of the formulated affinity matrix usually correspond to different latent objects in an image, the proposed method can be applied to various computer vision tasks, such as saliency detection, object proposal generation, and so on.

The remainder of this paper is organized as follows. In Section 2, we review some existing spectral clustering methods for image segmentation. In Section 3, we detail the construction procedure of the affinity matrix and the proposed method for object discovery. Moreover, we show the advantages of the proposed method over several existing spectral clustering methods for image segmentation. In Section 4, we apply the proposed method to saliency detection and object proposal generation. In Section 5, we evaluate the proposed method and present experimental results on several challenging benchmark datasets. Finally, we draw conclusions in Section 6.
\section{Related Work}
An affinity matrix plays an important role in the spectral graph theory~\cite{biggs1993}.
For example, Shi et al.~\cite{Shi2000} propose to solve the perceptual grouping problem using normalized cuts. They set up a weighted graph and set the weight of each edge connecting two nodes to be a measure of the affinity between the two nodes. Then image segmentation is treated as a graph partitioning problem, which is to find the minimum cut of a graph by solving a generalized eigenvalue problem on the constructed Laplacian matrix. The second-to-last eigenvectors turn out to be indicator vectors for partitioning the graph. Perona et al.~\cite{Freeman1998} consider an asymmetric variant of the cost function in~\cite{Shi2000}, and define one of the two subsets of a graph to be the foreground and define its complement to be the background. Based on the modified normalized cuts, they derive a foreground cut method by performing affinity factorization on the affinity matrix. The first eigenvector of the affinity matrix with the largest eigenvalue contains salient objects.

The application of the proposed CM-based method on saliency detection is similar to the above-mentioned foreground cut method. The first several eigenvectors of the proposed affinity matrix ranked by its  eigenvalues in descending order can be used to detect salient objects. However, those methods aforementioned (i.e.,~\cite{Shi2000} and ~\cite{Freeman1998}) do not consider color distortions in the affinity matrices. Image pixels in a small patch of an image, sharing the same real-world colors, may be similar to each other, while image pixels across the whole object may be quite different because of color distortions. Therefore, instead of defining an affinity matrix for all pairwise pixels in an image, we define an affinity matrix based on local affinity of image pixels. LPP~\cite{LPP2003} builds a graph incorporating the neighborhood information of input data and computes an affinity matrix to map the input data points into a subspace. Although LPP computes an affinity matrix based on a small local window, it does not consider color distortions. Several other spectral clustering methods (e.g.,~\cite{CaiC15,YangC15}) do not consider color distortions as well. Levin et al.~\cite{Levin2006} formulate an affinity matrix based on color lines for the task of natural image matting, where they use the affinity matrix to construct the Laplacian matrix. The eigenvector corresponding to the smallest eigenvalue of the Laplacian matrix contains a high-quality matte for the task of natural image matting. In this paper, we define a similar affinity matrix based on color lines and explore the eigenvectors of the affinity matrix (instead of the eigenvectors of the Laplacian matrix) for the task of object discovery. Moreover, we also analyze the theoretical superiority of the proposed method over several existing spectral clustering based methods for object discovery.
\section{Object Discovery}
In the proposed object discovery method, we formulate a novel affinity matrix, which is robust to color distortions. The affinity values among image pixels can be either positive or negative. Usually, positive affinity values are obtained if the pixels are similar and spatially adjacent. Otherwise, non-positive affinity values are obtained. Then, image pixels are clustered by using the built affinity matrix. During the process of clustering image pixels, a new measure for evaluating the goodness of clusters, which we call as cohesion measurement, is derived. The cohesion value of a cluster becomes large when the number of similar and spatially adjacent pixels grows. While the cohesion value decreases when the pixels of a cluster have more different values. Such differences are often caused by strong edges or the contrast of a salient object over background. Thus, an object, which has image pixels with similar real-world color and compact spatial positions, often obtains a high cohesion value. In this section, the details of the proposed method are given on how this cohesion measurement, which is encoded in the affinity matrix, is used for the task of object discovery.
\subsection{The Affinity Matrix Based on Color Distortions}
\label{sec:les}
Due to color distortions, image pixels on an object may share similar color within its small neighboring region. However, image pixels could be quite different from each other across the whole object. To address this issue, we propose an affinity matrix based on local affinity of image pixels. The proposed affinity matrix is inspired by the matting Laplacian matrix defined in~\cite{Levin2006}. Given a pixel $p_i(x, y)$ in an image of $w{\times}h$, where $w$ and $h$ are respectively the width and height of the image, according to~\cite{Levin2006}, the entry of the matting Laplacian matrix $L$ at $(i,j)$ is written as:
\begin{equation}
\begin{small}
\sum_{k|(i,j)\in\Omega_k}{(\delta_{ij}-\frac{1+(p_i-\mu_k)^T (\Sigma_k+\frac{\tau}{|\Omega_k|} E)^{-1} (p_j-\mu_k)}{|\Omega_k|})},
\end{small}
\label{eq::mattingLaplacian}
\end{equation}
where $k=x\cdot w+y$, and it denotes the index of each window $\Omega_k$ surrounding the pixel at $(x, y)$; $|\Omega_k|$ denotes the number of pixels in the window; $\mu_k$ and $\Sigma_k$ are the mean vector and covariance matrix of the pixel values in the window $\Omega_k$, respectively; $\delta_{ij}$ is the Kronecker delta. In order to avoid division by zero, a small constant value $\tau$ and an identity matrix $E$ are introduced to Eqn~(\ref{eq::mattingLaplacian}).

An affinity matrix $A^M$ can be derived from the matting Laplacian matrix by rewriting $A^M=D^M-L$, where $D^M$ is a diagonal matrix with each diagonal element $D^M_{ii}=\sum_{j}{A^M_{ij}}$. The element of the affinity matrix $A^M$ at $(i, j)$ is given by Eqn.~(\ref{eq::mattingAffinity}).

\begin{equation}
\begin{small}
A^M_{ij}=\sum_{k|(i,j)\in\Omega_k}{\frac{(1+(p_i-\mu_k)^T (\Sigma_k+\frac{\tau}{|\Omega_k|} E)^{-1} (p_j-\mu_k))}{|\Omega_k|}}.
\end{small}
\label{eq::mattingAffinity}
\end{equation}
Such affinity matrix is robust to color distortions. For simplification, we discard the scale factor $\frac{1}{|\Omega_k|}$ (which is a constant for a given window) and rewrite the affinity matrix as Eqn.~(\ref{eq::eqvAffinity}).
\begin{equation}\label{eq::eqvAffinity}
\begin{small}
  A_{ij}=\frac{\sum_{k|(i,j)\in\Omega_k}{(1+(p_i-\mu_k)^T (\Sigma_k+\tau E)^{-1} (p_j-\mu_k))}}{\sqrt{D_{ii}D_{jj}}},
  \end{small}
\end{equation}
where $D_{ii}=\sum_{j}{A_{ij}}$ and $D_{jj}=\sum_{i}{A_{ij}}$.
$A_{ij}$ keeps the average affinity between any two pixels in an image and the denominator in Eqn.~(\ref{eq::eqvAffinity}) averages the affinity value. Inspired by the mean filter~\cite{Gonzalez2006}, we expect that the average affinity is robust to noises. The main part of the numerator in Eqn.~(\ref{eq::eqvAffinity}) gives an affinity between pixel $p_i$ and $p_j$, which is weighted by color component. For convenience, we define the main part of the numerator in Eqn.~(\ref{eq::eqvAffinity}) as Eqn.~(\ref{eq::eqAE}).
\begin{equation}\label{eq::eqAE}
\begin{small}
  A_E=(p_i-\mu_k)^T (\Sigma_k+\tau E)^{-1} (p_j-\mu_k).
  \end{small}
\end{equation}
$A_E$ is robust to variations in lighting and shadows, which will be explained in detail next. Substituting Eqn.~(\ref{eq::eqAE}) to Eqn.~(\ref{eq::eqvAffinity}), we have
\begin{equation}\label{eq::eqAEAffinity}
\begin{small}
  A_{ij}=\frac{\sum_{k|(i,j)\in\Omega_k}{(1+A_E)}}{\sqrt{D_{ii}D_{jj}}}.
\end{small}
\end{equation}

Let $U$ be the matrix where each column is the eigenvector of the covariance matrix $\Sigma_k$, and let $\phi_z$ be the $z$th eigenvalue of the covariance matrix $\Sigma_k$. Then,
\begin{equation}
 A_E=\mathbf{\theta} \cdot (U(p_i-\mu_k))^T (U(p_j-\mu_k)),
 \label{innerProduct}
\end{equation}
 where $\mathbf{\theta}_z=\frac{1}{\phi_z+\tau}$ ($\theta_z>0$) is the $z$th element of the weight vector $\mathbf{\theta}$.
The derivation of Eqn.~(\ref{innerProduct}) can be seen in \textbf{Appendix A}. $(U(p_i-\mu_k))^T (U(p_j-\mu_k))$ is the inner product between the vector $(p_i-\mu_k)$ and the vector $(p_j-\mu_k)$ in the reference frame whose basis vectors are the columns of the matrix $U$. Since an orthogonal transformation and a vector translation do not change the inner product, $(U(p_i-\mu_k))^T (U(p_j-\mu_k))$ is also the inner product of the image pixels $p_i$ and $p_j$, which defines the affinity between $p_i$ and $p_j$.
In addition, the $z$th element $\theta_z$ of the weight vector $\mathbf{\theta}$ is nearly inversely proportional to the eigenvalue $\phi_z$. The weight vector $\theta$ alleviates the problem that a certain color component with a large variance dominates an affinity value, by which the obtained affinity value is robust to color distortions. For example, the saturations of a certain color component of image pixels in a small patch may be quite different because of color distortions. The resulting affinity values obtained by using the inner product based on these image pixels may be quite small but they have essentially a similar color. Fortunately, the certain color component $z$ with a large variance (i.e., a principle component) corresponds to a large eigenvalue $\phi_z$ and the corresponding weight value is small. Thus, $A_E$ is robust to color distortions for representing similarities between image pixels. An illustration of $A_E$ can be found in \textbf{Appendix B}. $A_{ij}$, which is the mean value of ($A_E+1$) over a small window (see Eqn.~(\ref{eq::eqAEAffinity})), is also robust to color distortions.
$A_{ij}$ maintains the average affinity between any two pixels over multiple windows in an image.
Usually, positive affinity values are obtained if image pixels in a window share a similar real-world color. Otherwise, non-positive affinity values are obtained.

\subsection{Clustering Using the Affinity Matrix}
\begin{figure*}[t]
\begin{center}
   \subfigure[]{
   \begin{minipage}[b]{0.133\textwidth}
   \includegraphics[width=1\textwidth]{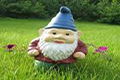}\\
   \includegraphics[width=1\textwidth]{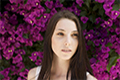}
   \end{minipage}}
   \subfigure[]{
   \begin{minipage}[b]{0.133\textwidth}
   \includegraphics[width=1\textwidth]{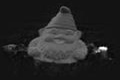}\\
   \includegraphics[width=1\textwidth]{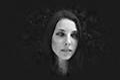}
   \end{minipage}}
   \subfigure[]{
   \begin{minipage}[b]{0.133\textwidth}
   \includegraphics[width=1\textwidth]{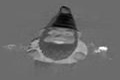}\\
   \includegraphics[width=1\textwidth]{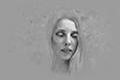}
   \end{minipage}}
    \subfigure[]{
   \begin{minipage}[b]{0.133\textwidth}
   \includegraphics[width=1\textwidth]{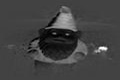}\\
   \includegraphics[width=1\textwidth]{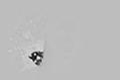}
   \end{minipage}}
    \subfigure[]{
   \begin{minipage}[b]{0.133\textwidth}
   \includegraphics[width=1\textwidth]{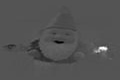}\\
   \includegraphics[width=1\textwidth]{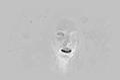}
   \end{minipage}}
    \subfigure[]{
   \begin{minipage}[b]{0.133\textwidth}
   \includegraphics[width=1\textwidth]{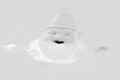}\\
   \includegraphics[width=1\textwidth]{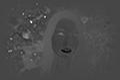}
   \end{minipage}}
    \subfigure[]{
   \begin{minipage}[b]{0.133\textwidth}
   \includegraphics[width=1\textwidth]{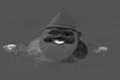}\\
    \includegraphics[width=1\textwidth]{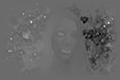}
   \end{minipage}}
\end{center}
   \vspace{-0.251cm}
   \caption{\label{eigenVectorFig}The eigenvectors of the affinity matrix contain the information of objects. (a) The input images. (b) to (g) The object maps corresponding to the first six eigenvectors ranked by their eigenvalues in descending order.}
\end{figure*}
As shown in Eqn.~(\ref{eq::eqvAffinity}), the affinity matrix $A$ is symmetric and it is a local representation for each pixel $p_i$ within a small window $\Omega_k$ surrounding $p_i$. Thus, the matrix $A$ is very sparse.
The affinity matrix $A$ keeps the pairwise affinity between pixels, which is robust to color distortions caused by variations in lighting and shadows. Thus, we can cluster the pixels, which are spatially adjacent and share similar real-world colors, by using the affinity matrix $A$. The aim of clustering is to verify whether pairwise pixels in a small window $\Omega_k$ belong to the same cluster or not, which can be written in Eqn.~(\ref{eq::cluster}).
\begin{equation}\label{eq::cluster}
  M_{ij}= \left\{
       \begin{array}{ll}
       1 & A_{ij}> \Delta , k|(i,j)\in\Omega_k \\
       -1 & \mbox{Otherwise}
       \end{array}
       \right.,
\end{equation}
where $\Delta$ is a threshold, which is used to partition adjacent pixels. Eqn.~(\ref{eq::cluster}) converts a clustering problem into a labeling problem. Typically, $\Delta$ is set to a constant value to obtain $M$. As a matter of fact, the optimal $M$ can be found by solving an optimization problem instead of setting a constant value to $\Delta$. Assume that we have an oracle $\tilde{M}$, a measure of quality can be obtained by measuring the Pearson correlation coefficient between the affinity matrix $A$ and the oracle $\tilde{M}$. To measure this correlation, we use the kernel-target alignment function proposed in \cite{Cristianini02}:
\begin{equation}\label{Eq::clusterCost}
  f(K_1,K_2)=\frac{\langle K_1,K_2\rangle_F}{\sqrt{\langle K_1,K_1\rangle_F \langle K_2,K_2\rangle_F}},
\end{equation}
where $K_d$ ($d$=1 or 2) is the $d$th kernel matrix for the image pixels, $\langle K_1,K_2\rangle_F=\sum_{i,j=1}^N{K_1(p_i,p_j)K_2(p_i,p_j)}$ and $F$ denotes the Frobenius norm. The kernel-target alignment can also be viewed as the \emph{Cosine} of the angle between two bi-dimensional vectors $K_1$ and $K_2$. In this paper, $N$ denotes the number of image pixels. Estimating an optimized oracle $\hat{M}$ is equivalent to maximizing the alignment between $A$ and the oracle $\tilde{M}$, which is modelled as:
\begin{equation}\label{Eq::clusterCost}
  \hat{M}=\mathop{\argmax}_{\tilde{M}}{f(A,\tilde{M})}=\mathop{\argmax}_{\tilde{M}}{\frac{\langle A,\tilde{M}\rangle_F}{\sqrt{\langle A,A\rangle_F \langle \tilde{M},\tilde{M}\rangle_F}}}.
\end{equation}
The kernel-target alignment between $A$ and the oracle $\tilde{M}$ measures how the pixels of the same cluster are close to each other, and it also measures how the pixels of different clusters are far away from each other at the same time.
 If we consider $\tilde{M}=\mathbf{\ell}\mathbf{\ell}^T$, where $\mathbf{\ell}$ is a $N$ dimensional vector whose elements belong to $\{-1,+1\}$, then
\begin{equation}\label{Eq::clusterCost2}
 f(A,\mathbf{\ell}\mathbf{\ell}^T)=\frac{\langle A,\mathbf{\ell}\mathbf{\ell}^T\rangle_F}{\sqrt{\langle A,A\rangle_F \langle \mathbf{\ell}\mathbf{\ell}^T,\mathbf{\ell}\mathbf{\ell}^T\rangle_F}}=\frac{\langle A,\mathbf{\ell}\mathbf{\ell}^T\rangle_F}{\langle A\rangle_F \langle \ell\ell^T\rangle_F}.
\end{equation}
Since $A$ is a constant Hermitian matrix once an image is given, multiplying $f(A,\mathbf{\ell}\mathbf{\ell}^T)$ by a constant $\langle A\rangle_F$ will not change the solution of maximizing the alignment $f(A,\mathbf{\ell}\mathbf{\ell}^T)$. Therefore, equivalently, the solution of Eqn.~(\ref{Eq::clusterCost}) can be obtained by solving Eqn.~(\ref{Eq::clusterCost3}).
\begin{equation}\label{Eq::clusterCost3}
  \mathbf{\ell}=\mathop{\argmax}_\mathbf{\ell}{f(A,\mathbf{\ell}\mathbf{\ell}^T)}=\mathop{\argmax}_\mathbf{\ell \neq 0}{\frac{\mathbf{\ell}^TA\mathbf{\ell}}{\mathbf{\ell}^T \mathbf{\ell}}}.
\end{equation}
According to the Rayleigh-Ritz theorem~\cite{Horn1986}, the optimal solution of Eqn.~(\ref{Eq::clusterCost3}) can be approximated by the first eigenvector of the affinity matrix $A$, which is written as Eqn.~(\ref{Eq::RayleighTheorem}), Eqn.~(\ref{Eq::RayleighTheorem2}) and Eqn.~(\ref{Eq::RayleighTheorem3}).
\begin{equation}\label{Eq::RayleighTheorem}
  \lambda_1 \mathbf{\ell}^T \mathbf{\ell} \geq \mathbf{\ell}^TA\mathbf{\ell} \geq \lambda_N \mathbf{\ell}^T \mathbf{\ell},
\end{equation}
where $\lambda_i$ is the $i$th eigenvalue of the affinity matrix $A$.
\begin{equation}\label{Eq::RayleighTheorem2}
  \lambda_{max}=\lambda_1=\mathop{\max}_{\mathbf{\ell} \neq 0}{\frac{\mathbf{\ell}^TA\mathbf{\ell}}{\mathbf{\ell}^T \mathbf{\ell}}}=\mathop{\max}_{\mathbf{\ell}^T \mathbf{\ell}=1}{\mathbf{\ell}^T A\mathbf{\ell}}.
\end{equation}
\begin{equation}\label{Eq::RayleighTheorem3}
  \lambda_{min}=\lambda_N=\mathop{\min}_{\mathbf{\ell} \neq 0}{\frac{\mathbf{\ell}^TA\mathbf{\ell}}{\mathbf{\ell}^T \mathbf{\ell}}}=\mathop{\min}_{\mathbf{\ell}^T \mathbf{\ell}=1}{\mathbf{\ell}^TA\mathbf{\ell}}.
\end{equation}
The theorem also shows that the maximum of the alignment $f(A,\mathbf{\ell}\mathbf{\ell}^T)$ is equal to the largest eigenvalue of the affinity matrix $A$. The bigger the eigenvalue is, the bigger the alignment value is. The eigenvector corresponding to each eigenvalue is the obtained local maximum solution of Eqn.~(\ref{Eq::clusterCost3}), which is a soft continuous solution (i.e., continuous labels for image pixels). As the consequence, the total number of clusters is equal to the number of eigenvectors.

Usually, different clusters (i.e., different object regions) obtained from different eigenvectors correspond to different eigenvalues. A salient object, which has high contrast relative to its background, corresponds to a higher eigenvalue. These cases will be explained in detail in the following subsection.
\subsection{Cohesion Measurement for Object Regions}
In this paper, the concept of cohesion is adopted to measure the tightness of a cluster. Cluster cohesion is defined as the sum of the weights of all the edges between the nodes within a cluster. Another term is cluster separation, which is defined as the sum of the weights of the edges between the nodes from different clusters. Usually, cluster cohesion and cluster separation measure intra-class affinity and inter-class affinity, respectively. Different clustering results lead to different cluster cohesion and cluster separation values. In our case, the cohesion value of a cluster is obtained by performing eigenvalue decomposition on the affinity matrix. As a result, the eigenvalue corresponds to the degree of the cohesion of one cluster. The higher the cohesion value is, the tighter the cluster is. Basically, the eigenvalues of the affinity matrix measure the quality of clusters. Moreover, we expect that different clusters correspond to different object regions. Thus, the eigenvalues of the affinity matrix are treated as the degrees of cohesion for object regions.
In addition to being capable of identifying object regions in an image, the cohesion measurement can also be used to find salient objects. Salient object regions in an image, which have high contrast relative to the background, correspond to higher cohesion values.

To further show the relationship between the degrees of cohesion and object regions, the elements of the label vector $\ell$ for image pixels obtained from Eqn.~(\ref{Eq::clusterCost3}) are analyzed in two cases, which are listed in Eqn.~(\ref{eq::labelDiscussion}).
\begin{equation}\label{eq::labelDiscussion}
\begin{small}
 \mathop{\max}_{\ell^T\ell = 1}{\sum_{ij}{{\ell_i}A_{ij}{\ell_j}}}\Rightarrow \\
 \left\{
       \begin{array}{ll}
        sgn(\ell_i)=sgn(\ell_j), &A_{ij}\geq 0 \\
        sgn(\ell_i)\neq sgn(\ell_j), &A_{ij}\leq 0 \\
       \end{array}
       \right.,
 \end{small}
\end{equation}
where $sgn(\cdot)$ is the sign function that returns the sign of a variable. As shown in Eqn.~(\ref{eq::labelDiscussion}), in order to maximize the objective function, $\ell_i$ and $\ell_j$ should have the same sign when $A_{ij}$ is positive. Otherwise, $\ell_i$ and $\ell_j$ should have different signs when $A_{ij}$ is negative. In both cases, the maximum of the objective function is obtained when both $A_{ij}$ and $\ell_i\ell_j$ are large positive values or small negative values. This in turn implies that ${\ell_i}A_{ij}{\ell_j}$ has a big value. When $A_{ij}$ and $\ell_i\ell_j$ are large positive values, image regions are obtained by checking the signs of labels for spatial adjacent image pixels. In the other case (i.e., $A_{ij}$ and $\ell_i\ell_j$ are small negative values), the labeling values $\ell_i$ and $\ell_j$ should be far away from each other under the constraint of $\ell^T\ell = 1$.
Usually, negative affinity values are caused by different pixel values, which are due to the factors such as color distortions, color contrast, strong edges, etc.
However, the formulated affinity matrix is robust to color distortions, which are caused by variations in lighting and shadows. As a result, the difference between image pixels due to color distortions will not lead to a big cohesion value. Therefore, the big cohesion value is mainly due to a large number of spatially adjacent similar pixels, high contrast relative to background regions and strong edges, which imply a salient object region.

\subsection{Object Discovery Using the Affinity Matrix}
\begin{figure}
   \begin{center}
   \subfigure{
   \begin{minipage}[b]{0.14\textwidth}
   \includegraphics[width=1\textwidth]{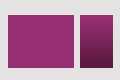}
   \vspace{-0.7cm}
   \caption*{\scriptsize(a1)}
   \end{minipage}}
   \subfigure{
   \begin{minipage}[b]{0.14\textwidth}
   \includegraphics[width=1\textwidth]{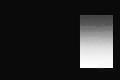}
   \vspace{-0.7cm}
   \caption*{\scriptsize(b1)}
   \end{minipage}}
   \subfigure{
   \begin{minipage}[b]{0.14\textwidth}
   \includegraphics[width=1\textwidth]{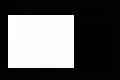}
   \vspace{-0.7cm}
   \caption*{\scriptsize(c1)}
   \end{minipage}}
   \subfigure{
   \vspace{-0.7cm}
   \begin{minipage}[b]{0.14\textwidth}
   \includegraphics[width=1\textwidth]{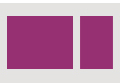}
   \vspace{-0.7cm}
   \caption*{\scriptsize(a2)}
   \end{minipage}}
   \subfigure{
   \vspace{-0.7cm}
   \begin{minipage}[b]{0.14\textwidth}
   \includegraphics[width=1\textwidth]{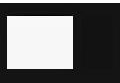}
   \vspace{-0.7cm}
   \caption*{\scriptsize(b2)}
   \end{minipage}}
   \subfigure{
   \vspace{-0.7cm}
   \begin{minipage}[b]{0.14\textwidth}
   \includegraphics[width=1\textwidth]{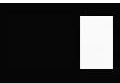}
   \vspace{-0.7cm}
   \caption*{\scriptsize(c2)}
   \end{minipage}}
\end{center}
   \vspace{-0.5cm}
   \caption{\label{Fig::SyntheticPrinciple}The demonstration of the ability that the proposed method can discover objects no matter that whether the objects undergo color distortions or not. (a1)/(a2) The two input images. (b1/b2) and (c1/c2) The object maps corresponding to the first two eigenvectors ranked by their eigenvalues in descending order.}
\end{figure}
As discussed in Section~\ref{sec:les}, because the affinity matrix is robust to color distortions, the clusters of pixels, which share a similar real-world color and spatially adjacency, are obtained by calculating the eigenvectors of the affinity matrix. Different clusters correspond to different object regions. In other words, the object regions can be segmented by using eigen-decomposition on the affinity matrix,
which is written as
\begin{equation}\label{eigenVector}
  A \mathbf{\nu}=\lambda \mathbf{\nu},
\end{equation}
where $\mathbf{\nu}$ and $\lambda$ are the eigenvectors and eigenvalues of $A$, respectively.

Generally speaking, one resulting eigenvector corresponds to one latent object region in an image, although it may be affected by some noises. One element in an eigenvector corresponds to one pixel in the image. As a result, it becomes convenient to relate the eigenvectors to the object regions in an image. Given that $\nu_k$ is the $k$th element of an eigenvector $\nu$, the pixel value at $(x,y)$ in the object map derived from $\nu_k$ and $\nu$ is written as
\begin{equation}\label{eqn:norm}
V(x,y)=\frac{\mathbf{\nu}_k-min(\mathbf{\nu})}{max(\mathbf{\nu})-min(\mathbf{\nu})}{\cdot}C,
\end{equation}
where $x=\frac{k}{w}$, $y=mod(k,w)$ and $C$ denotes a contant.
Fig.~\ref{eigenVectorFig} shows the object maps derived from the first six eigenvectors corresponding to the six largest eigenvalues obtained by Eqn.~(\ref{eqn:norm}). As shown in Fig.~\ref{eigenVectorFig}, several salient objects or object regions are found according to the first several eigenvectors. Moreover, the proposed method also enumerates object regions at different scales corresponding to different cohesion values.

\begin{figure}
   \begin{center}
   \subfigure{
   \begin{minipage}[b]{0.14\textwidth}
   \includegraphics[width=1\textwidth]{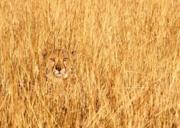}
   \vspace{-0.7cm}
   \caption*{\scriptsize(a1)}
   \end{minipage}}
   \subfigure{
   \begin{minipage}[b]{0.14\textwidth}
   \includegraphics[width=1\textwidth]{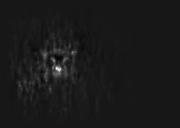}
   \vspace{-0.7cm}
   \caption*{\scriptsize(b1)}
   \end{minipage}}
   \subfigure{
   \begin{minipage}[b]{0.14\textwidth}
   \includegraphics[width=1\textwidth]{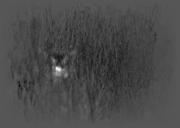}
   \vspace{-0.7cm}
   \caption*{\scriptsize(c1)}
   \end{minipage}}
   \subfigure{
   \vspace{-0.7cm}
   \begin{minipage}[b]{0.14\textwidth}
   \includegraphics[width=1\textwidth]{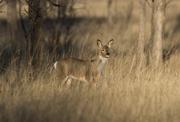}
   \vspace{-0.7cm}
   \caption*{\scriptsize(a2)}
   \end{minipage}}
   \subfigure{
   \vspace{-0.7cm}
   \begin{minipage}[b]{0.14\textwidth}
   \includegraphics[width=1\textwidth]{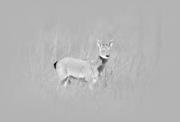}
   \vspace{-0.7cm}
   \caption*{\scriptsize(b2)}
   \end{minipage}}
   \subfigure{
   \vspace{-0.7cm}
   \begin{minipage}[b]{0.14\textwidth}
   \includegraphics[width=1\textwidth]{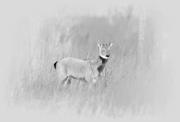}
   \vspace{-0.7cm}
   \caption*{\scriptsize(c2)}
   \end{minipage}}
\end{center}
   \vspace{-0.5cm}
   \caption{\label{Fig::SimilarBackground}The demonstration of the ability that the proposed method can discover objects even though the objects have similar colors to their respective background. (a1)/(a2) The two input images. (b1/b2) and (c1/c2) The object maps corresponding to the first two eigenvectors ranked by their eigenvalues in descending order.}
\end{figure}
\begin{figure}
\begin{center}
   \subfigure[]{\includegraphics[width=0.15\textwidth]{face.jpg}}
   \subfigure[]{\includegraphics[width=0.15\textwidth]{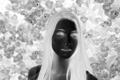}}
   \subfigure[]{\includegraphics[width=0.15\textwidth]{face__1.jpg}}
\end{center}
   \vspace{-0.251cm}
   \caption{\label{Fig:CompEigenvector}The object maps derived from the eigenvectors of the proposed affinity matrix and those of the matting Laplacian matrix. (a) The input Image. (b) The object map derived from the last eigenvector (with the smallest eigenvalue) of the matting Laplacian matrix as defined in~\cite{Levin2006}. Note that the last eigenvector of the matting Laplacian matrix is the best matting eigenvector as described in~\cite{Levin2006}. (c) The object map derived from the first ranked eigenvector (with the largest eigenvalue) of the proposed affinity matrix.}
\end{figure}
\begin{figure*}
\begin{center}
   \subfigure[]{
   \begin{minipage}[b]{0.13\textwidth}
   \includegraphics[width=1\textwidth]{face.jpg}\\
   \includegraphics[width=1\textwidth]{face.jpg}\\
   \includegraphics[width=1\textwidth]{face.jpg}
   \end{minipage}}
   \subfigure[]{
   \begin{minipage}[b]{0.13\textwidth}
   \includegraphics[width=1\textwidth]{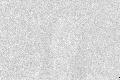}\\
   \includegraphics[width=1\textwidth]{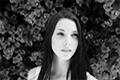}\\
   \includegraphics[width=1\textwidth]{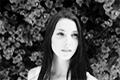}
   \end{minipage}}
   \subfigure[]{
   \begin{minipage}[b]{0.13\textwidth}
   \includegraphics[width=1\textwidth]{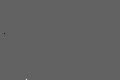}\\
   \includegraphics[width=1\textwidth]{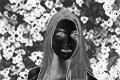}\\
   \includegraphics[width=1\textwidth]{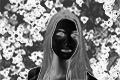}
   \end{minipage}}
   \subfigure[]{
   \begin{minipage}[b]{0.13\textwidth}
   \includegraphics[width=1\textwidth]{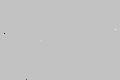}\\
   \includegraphics[width=1\textwidth]{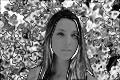}\\
   \includegraphics[width=1\textwidth]{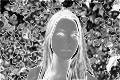}
   \end{minipage}}
   \subfigure[]{
   \begin{minipage}[b]{0.13\textwidth}
   \includegraphics[width=1\textwidth]{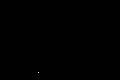}\\
   \includegraphics[width=1\textwidth]{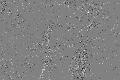}\\
   \includegraphics[width=1\textwidth]{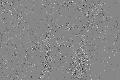}
   \end{minipage}}
   \subfigure[]{
   \begin{minipage}[b]{0.13\textwidth}
   \includegraphics[width=1\textwidth]{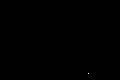}\\
   \includegraphics[width=1\textwidth]{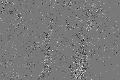}\\
   \includegraphics[width=1\textwidth]{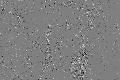}
   \end{minipage}}
   \subfigure[]{
   \begin{minipage}[b]{0.13\textwidth}
   \includegraphics[width=1\textwidth]{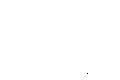}\\
   \includegraphics[width=1\textwidth]{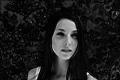}\\
   \includegraphics[width=1\textwidth]{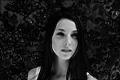}
   \end{minipage}}
\end{center}
   \vspace{-0.251cm}
   \caption{\label{Fig::compEigen}Examples of an input image and the object maps derived from the resulting eigenvectors of the Laplacian matrix in~\cite{Shi2000}, the affinity matrix in~\cite{Freeman1998} and the normalized Laplacian matrix in~\cite{Ng01} are shown in the first, the second and the third row, respectively. (a) The input image. (b)-(d) The object maps derived from the first three ranked eigenvectors obtained by the three methods. (e)-(g) The object maps derived from the last three ranked eigenvectors obtained by the three methods. Note that the last eigenvector obtained by~\cite{Shi2000} in the first row is $\mathbf{1}$-vector, whose elements are all one, since it is the last ranked eigenvector of the Laplacian matrix.}
\end{figure*}
To verify that the proposed method is able to find the cohesive clusters for image pixels, no matter whether they undergo color distortions or not, we illustrate an example in Fig. \ref{Fig::SyntheticPrinciple}. We synthesize two input images, where each input image contains two rectangles (shown in Fig.~\ref{Fig::SyntheticPrinciple}(a1) and (a2)). The pixels in the small rectangle in  Fig.~\ref{Fig::SyntheticPrinciple}(a1) undergoes intensity variation while the pixels in the other rectangles have the same pixel values. The resulting eigenvectors obtained by the proposed method coincide with the conclusion. As shown in Fig. \ref{Fig::SyntheticPrinciple}(b1) and (c1), the eigenvectors obtained from the input image in Fig.~\ref{Fig::SyntheticPrinciple}(a1), which undergoes color distortions caused by intensity variation, show that the proposed method is robust to intensity variation for object discovery. In addition, the small rectangle in Fig.~\ref{Fig::SyntheticPrinciple}(a1), which undergoes color distortions, is obtained in the first eigenvector. This shows that the high differences of the pixels in an object region correspond a high cohesion value. For the input image in Fig.~\ref{Fig::SyntheticPrinciple}(a2), image pixels in the rectangles are almost identical. The rectangle derived from the first eigenvector (see Fig.~\ref{Fig::SyntheticPrinciple}(b2)) corresponds to the one which has the largest area of image pixels. This shows that the proposed method tends to obtain the cluster with the maximum cohesion value.

It is worthy pointing out that the proposed method can find not only a salient object that has high contrast relative to its background, but also an object that has similar colors to its background. As shown in Fig.~\ref{Fig::SimilarBackground}(a1/a2), the leopard in the first row and the deer in the second row respectively have low contrast relative to its background. Fig.~\ref{Fig::SimilarBackground}(b1/b2) and (c1/c2) show the object maps derived from the resulting eigenvectors obtained by the proposed method. As we can see, the proposed method can still locate both the objects from the resulting eigenvectors since the objects have contours which prevent the pixels inside and outside of the objects from being clustered together.
\subsection{Relation to Spectral Clustering}
We note that a similar spectral clustering method was presented in~\cite{Levin2006}, which solves an eigen-decomposition problem on the matting Laplacian matrix. Although the affinity matrix in the proposed method plays a similar role as the matting Laplacian matrix, the resulting eigenvectors obtained by the two different matrices are quite different (see Fig.~\ref{Fig:CompEigenvector} for an illustration). In Fig.~\ref{Fig:CompEigenvector}(b), with the Laplacian matrix defined in~\cite{Levin2006}, only sub-optimal clustering results are obtained since the eigenvectors of the matting Laplacian matrix are not valid clusters of image pixels in object regions when the image pixels suffer from color distortions. In contrast, as shown in Fig.~\ref{Fig:CompEigenvector}(c), the proposed method obtains better clustering results than the method in~\cite{Levin2006}.
The proposed method tends to obtain different cohesive clusters by solving an eigen-decomposition problem on the formulated affinity matrix. Because the proposed affinity matrix is robust to color distortions, it can be more effectively used to find cohesive clusters for image pixels than the affinity matrices defined in the other spectral clustering methods. For instance, Shi and Malik~\cite{Shi2000}, Andrew et al.~\cite{Ng01} and Perona et al.~\cite{Freeman1998} construct affinity matrices only based on the Euclidean distance of image pixels in the RGB space. Among which, \cite{Shi2000} and \cite{Ng01} respectively use the eigenvectors of the Laplacian matrix and the normalized Laplacian matrix derived from the affinity matrices to segment images; While~\cite{Freeman1998} tries to find foreground regions by calculating the singular value decomposition of the constructed affinity matrix.
In fact, the essence of the method in~\cite{Freeman1998} is also to use the eigenvectors of the affinity matrix for image segmentation. Fig.~\ref{Fig::compEigen} shows the object maps derived from the first three eigenvectors and the last three eigenvectors of the affinity matrices or the Laplacian matrix obtained in these three methods. As shown in Fig.~\ref{Fig::compEigen}, the eigenvectors of the affinity matrices and the Laplacian matrix, which do not consider color distortions, contain many noises. Moreover, LLP~\cite{LPP2003} also faces a similar problem on the task of image segmentation. In contrast, the clustering results obtained by the proposed method contain less noises (see Fig.~\ref{Fig:CompEigenvector}(c)).
\section{Applications of The Proposed Method for Object Discovery}
In this section, we apply the proposed method to saliency detection and object proposal generation, respectively. Because each object map may contain a part of an object, the way of combining several object maps is also investigated in this section. Furthermore, inspired by the saliency filter method~\cite{saliencyFilters2012}, a strategy is proposed to eliminate the noisy regions in the obtained object maps.
\subsection{Saliency Detection}
\textbf{Saliency Detection with a Single Eigenvector:}
Firstly, we apply the obtained object maps to saliency detection, which aims to segment a salient object. For an object map $V$ generated by Eqn.~(\ref{eqn:norm}), the pixel intensities of an object may be reversed (i.e., the pixel values of the salient objects are lower than those of the background). To overcome this problem, we adopt a strategy which is given by Eqn.~(\ref{checkBackground}).
\begin{equation}\label{checkBackground}
  \tilde{V_t}=|V_t-\eta|,
\end{equation}
where $t$ is the index of the object maps; $\eta$ denotes the mean value of the pixels in the hull of each object map $V_t$ whose elements are the outermost pixels of the object map.
 A simple way to obtain a binary mask of an object is to use a threshold on the pixel intensities of the obtained object map, which has also been adopted in \cite{FrequencySalient2009,ChengPAMI2014}. The thresholding method is based on a clip-level (or a threshold value) to turn a gray-scale image (i.e., the object map in our case) into a binary image. Then an image morphology algorithm can be used on the binary image to generate the mask of objects.

\textbf{Saliency Detection with Multiple Eigenvectors:}
In the above proposed saliency detection method, a salient object is detected by considering only one eigenvector of the affinity matrix, in which we expect that one cohesive object region corresponds to one object and the object region is exactly determined by one eigenvector. However, in practice, different eigenvectors may correspond to different parts of an object due to the influence of color variations in the object. Thus, we also propose a saliency detection method, which considers the contributions from multiple eigenvectors to saliency detection. The object map $V_S$ derived from a combination of $S$ eigenvectors is defined as:
\begin{equation}\label{eigenCombin}
  V_S=\frac{\sum_{t=1}^S{\tilde{V_t}}}{S}.
\end{equation}
 Once we obtain the combined object map $V_S$, the thresholding method is then used on $V_S$ for detecting salient objects. However, the combination of several eigenvectors may generate some noisy regions, which do not belong to salient objects. Next, we will describe a noise elimination strategy to eliminate the noisy regions.

 \textbf{The Noise Elimination Strategy for Saliency Detection:}
 \label{eliNoise}
Ideally, the combined object map obtained from the first several eigenvectors contains salient objects. However, some noisy regions may exist in the combined object map. These noisy regions usually have low intensity values and a fragmented distribution. In contrast, the object regions in the combined object map often have a compact spatial distribution and stand out saliently. Therefore, we eliminate the noisy regions using the saliency measure defined in the salient filter method~\cite{saliencyFilters2012}.

\begin{figure}
\begin{center}
   \begin{minipage}[b]{0.15\textwidth}
   \includegraphics[width=1\textwidth]{1_30_30895__1.jpg}\\
   \includegraphics[width=1\textwidth]{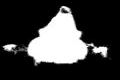}
   \end{minipage}
   \begin{minipage}[b]{0.15\textwidth}
   \includegraphics[width=1\textwidth]{1_30_30895__3.jpg}\\
   \includegraphics[width=1\textwidth]{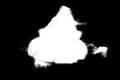}
   \end{minipage}
   \begin{minipage}[b]{0.15\textwidth}
   \includegraphics[width=1\textwidth]{1_30_30895__6.jpg}\\
    \includegraphics[width=1\textwidth]{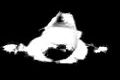}
   \end{minipage}
\end{center}
   \vspace{-0.251cm}
   \caption{\label{NoiseEli} Examples of saliency maps obtained by using the proposed noise elimination strategy on several object maps. From top to bottom, each row shows the object maps and the saliency maps, respectively.}
\end{figure}

In \cite{saliencyFilters2012}, an image is first abstracted into perceptually homogeneous elements using the superpixel segmentation method~\cite{SLIC2012}. Each element corresponds to a superpixel obtained by~\cite{SLIC2012}. Then, the element uniqueness (or contrast) and the element distribution of the homogeneous elements are defined to evaluate the saliency values of these elements. In their implementation, the mean color of each element is obtained as the representative color of the element.
 Usually, the mean colors of the elements obtained from an object region may be quite different from each other because of color distortions. However, the elements obtained from the combined object map are generally homogeneous. Therefore, the values of the element uniqueness and the element distribution of the segmented elements are calculated on the combined object map instead of the origin image in our implementation. Since the combined object map is a gray image, the mean intensity of each element is calculated in this task as the representative intensity of the element. Based on the element uniqueness and the element distribution, a saliency map is obtained by assigning a saliency value to each element obtained from a combined object map. The thresholding method will then be used on the obtained saliency map to segment salient objects. Fig.~(\ref{NoiseEli}) shows several examples of the resulting saliency maps obtained by using the proposed noise elimination strategy on several object maps. The noise regions in the object maps can be effectively eliminated when they have low intensity values and a fragmented distribution. However, some regions in a salient object may also be eliminated by the proposed strategy if they have similar pixel values to the background (see the last column in Fig.~(\ref{NoiseEli})).

\label{exp:saliency}
\begin{figure*}[t]
\begin{center}
   \subfigure[]{\includegraphics[width=0.32\textwidth]{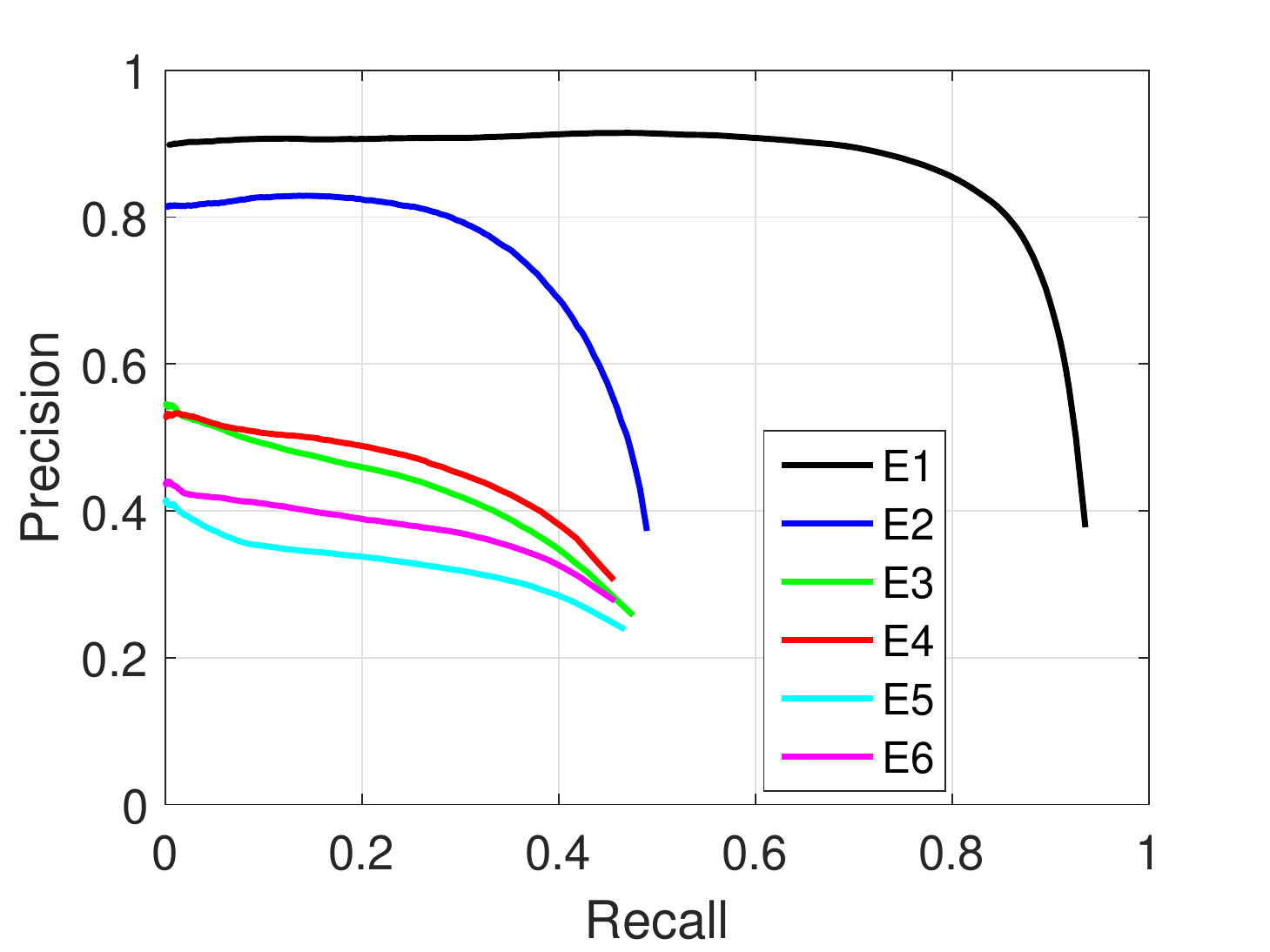}}
   \subfigure[]{\includegraphics[width=0.32\textwidth]{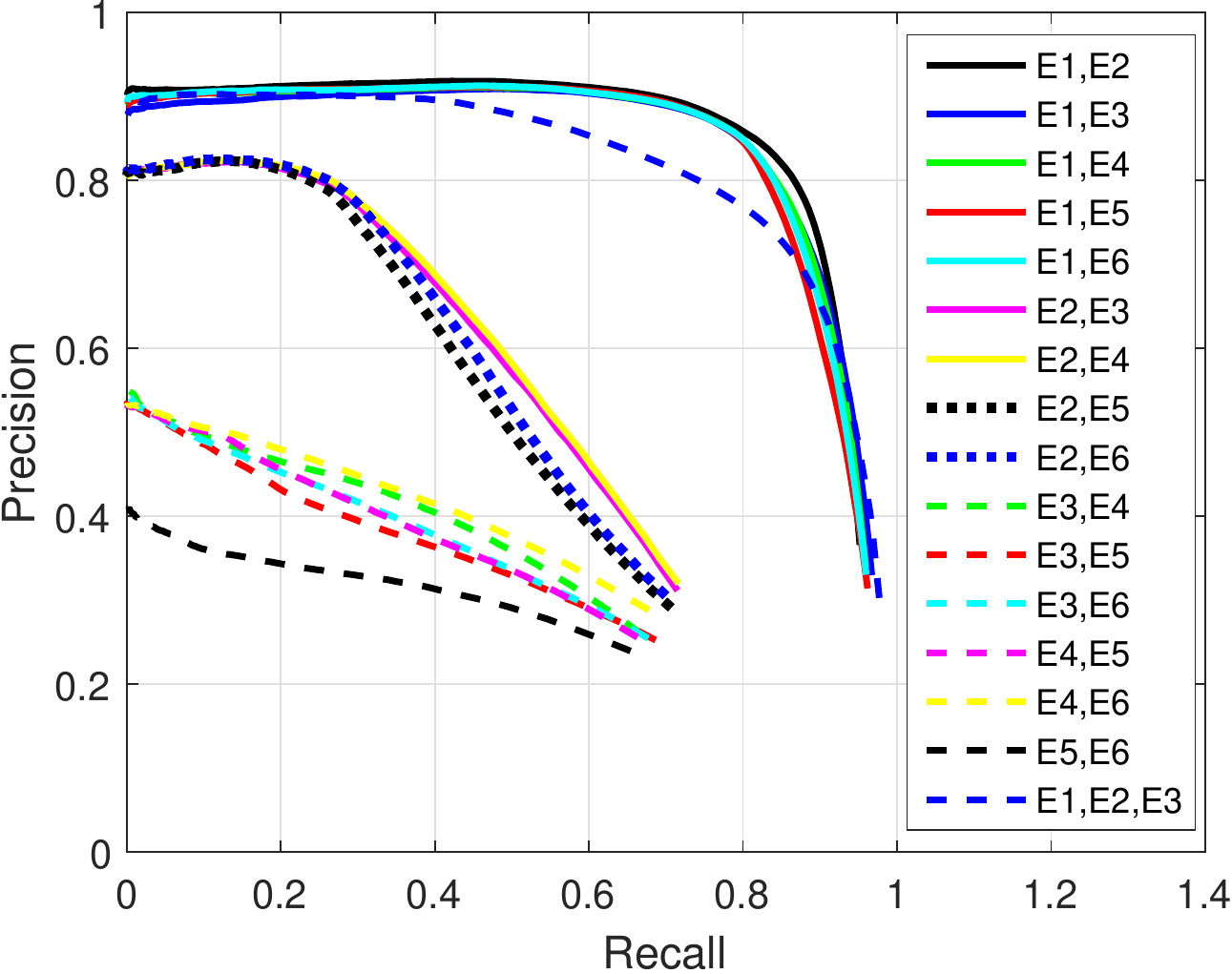}}
   \subfigure[]{\includegraphics[width=0.320\textwidth]{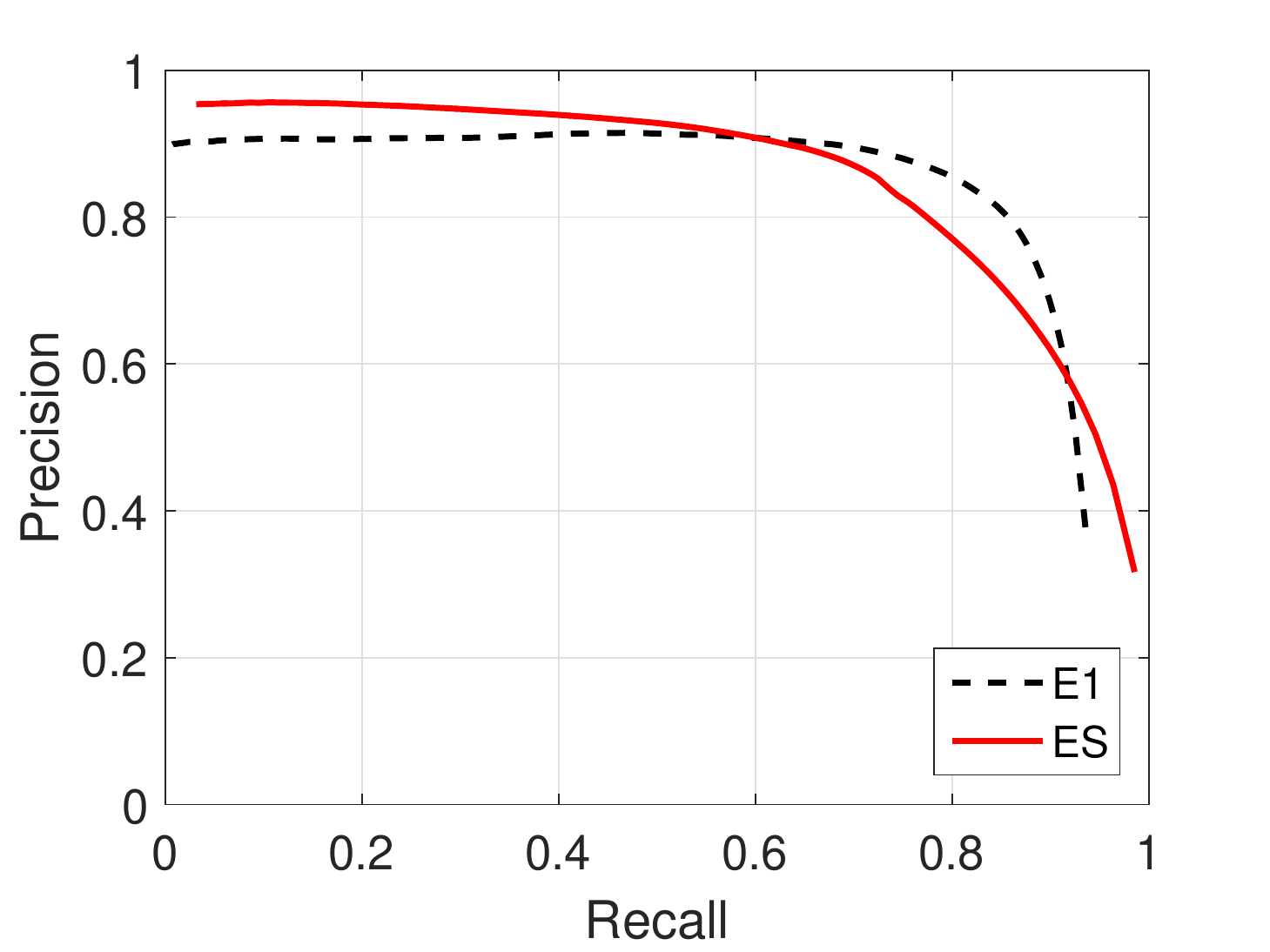}}
\end{center}
   \vspace{-0.251cm}
   \caption{\label{PRcurves}The PR curves obtained the proposed method on the MSRA10K dataset for saliency detection. (a) The results of saliency detection using one of the first six eigenvectors (E1 to E6) ranked by their eigenvalues, respectively. (b) The results of saliency detection using the combinations of the first six ranked eigenvectors. (c) Saliency detection using the first eigenvector only, and using the first eigenvector with the proposed noise elimination strategy for eliminating noisy regions (which is denoted by ES).}
\end{figure*}
 \subsection{Object Proposal Generation}
In this subsection, we apply the obtained object maps to the task of object proposal generation. In comparison to saliency detection, object proposal generation is more challenging because it is to recall all possible objects in an image. Inspired by the Edge Boxes~\cite{edgeBoxes2014} and gPb-owt-ucm~\cite{gPbowtucm2011} methods, we use the information from edges, which are a primitive feature of an image, to generate object proposals quickly. The Canny edge detector \cite{cannyEdge1986}, due to its simplicity and efficiency, is used to detect the edges in an object map. In order to obtain all possible object proposals, the Canny edge detector is performed on each chosen object map corresponding to an eigenvector of the affinity matrix. The number of eigenvectors used in the task of object proposal generation is more than that is used in the task of saliency detection. Similar to saliency detection, the combinations of the first several eigenvectors are also used to detect edges. All the circumscribed rectangles of the connected and closed edges are treated as the targeting object proposals.

As described in \cite{Objectness2012}, color contrast and closed boundary are two important objectness cues. Actually, object regions are obtained by using the pixel differences caused by color contrast and strong edges in the proposed method. The obtained edges from object maps are almost the significant edges in original images, while they may be mixed with a few insignificant edges obtained from noisy regions.  Therefore, the proposed method actually uses the two cues (i.e., color and edge) in a cascade way for the task of object proposal generation.

\textbf{Truncated Objectness Measure:}
Similar to the task of saliency detection, we use saliency measure to filter out noisy regions in the task of object proposal generation. Although saliency detection and objectness have different evaluation criteria, both of them aim to find object regions latent in an image. Salient objects should obtain high objectness measure scores, while background regions should obtain low saliency measure or objectness measure scores. Thus, we propose to use the saliency measure defined in~\cite{saliencyFilters2012} to measure the quality of the generated proposal windows. Perceptually homogeneous elements, which are obtained by using a superpixel segmentation method, are replaced by the proposal windows obtained from the object maps in this task. The saliency value of a proposal window, which does not hit any object region, is usually lower. Thus we call this measure as the truncated objectness measure.
\section{Experiments}
\begin{figure*}
\begin{center}
   \subfigure[Original]{
   \begin{minipage}[b]{0.080\textwidth}
   \includegraphics[width=1\textwidth]{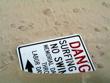}\\
   \includegraphics[width=1\textwidth]{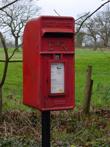}\\
   \includegraphics[width=1\textwidth]{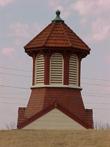}\\
   \includegraphics[width=1\textwidth]{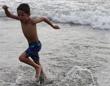}\\
   \includegraphics[width=1\textwidth]{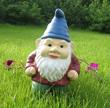}\\
   \includegraphics[width=1\textwidth]{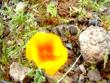}\\
   \includegraphics[width=1\textwidth]{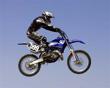}
   \end{minipage}}
   \subfigure[DSR]{
   \begin{minipage}[b]{0.080\textwidth}
   \includegraphics[width=1\textwidth]{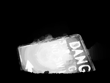}\\
   \includegraphics[width=1\textwidth]{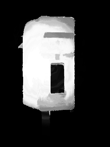}\\
   \includegraphics[width=1\textwidth]{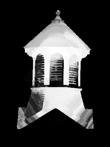}\\
   \includegraphics[width=1\textwidth]{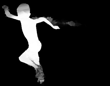}\\
   \includegraphics[width=1\textwidth]{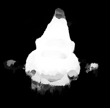}\\
   \includegraphics[width=1\textwidth]{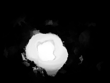}\\
   \includegraphics[width=1\textwidth]{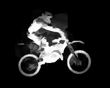}
   \end{minipage}}
   \subfigure[GR]{
   \begin{minipage}[b]{0.080\textwidth}
   \includegraphics[width=1\textwidth]{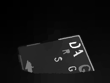}\\
   \includegraphics[width=1\textwidth]{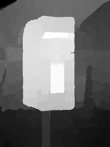}\\
   \includegraphics[width=1\textwidth]{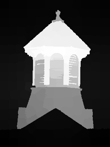}\\
   \includegraphics[width=1\textwidth]{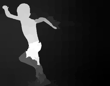}\\
   \includegraphics[width=1\textwidth]{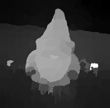}\\
   \includegraphics[width=1\textwidth]{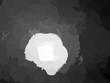}\\
   \includegraphics[width=1\textwidth]{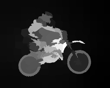}
   \end{minipage}}
   \subfigure[MC]{
   \begin{minipage}[b]{0.080\textwidth}
   \includegraphics[width=1\textwidth]{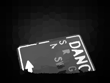}\\
   \includegraphics[width=1\textwidth]{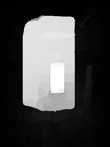}\\
   \includegraphics[width=1\textwidth]{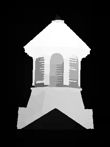}\\
   \includegraphics[width=1\textwidth]{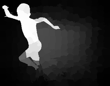}\\
   \includegraphics[width=1\textwidth]{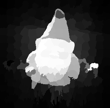}\\
   \includegraphics[width=1\textwidth]{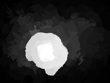}\\
   \includegraphics[width=1\textwidth]{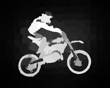}
   \end{minipage}}
   \subfigure[MNP]{
   \begin{minipage}[b]{0.080\textwidth}
   \includegraphics[width=1\textwidth]{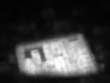}\\
   \includegraphics[width=1\textwidth]{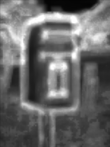}\\
   \includegraphics[width=1\textwidth]{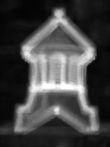}\\
   \includegraphics[width=1\textwidth]{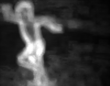}\\
   \includegraphics[width=1\textwidth]{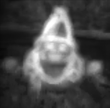}\\
   \includegraphics[width=1\textwidth]{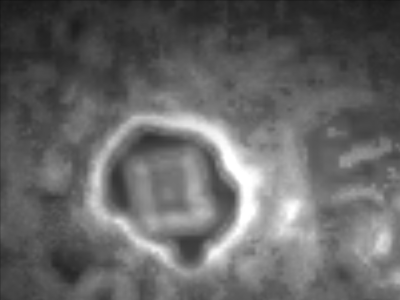}\\
   \includegraphics[width=1\textwidth]{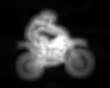}
   \end{minipage}}
   \subfigure[RC]{
   \begin{minipage}[b]{0.080\textwidth}
   \includegraphics[width=1\textwidth]{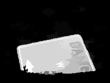}\\
   \includegraphics[width=1\textwidth]{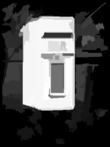}\\
   \includegraphics[width=1\textwidth]{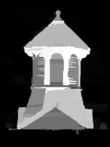}\\
   \includegraphics[width=1\textwidth]{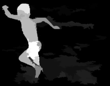}\\
   \includegraphics[width=1\textwidth]{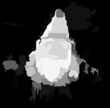}\\
   \includegraphics[width=1\textwidth]{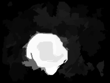}\\
   \includegraphics[width=1\textwidth]{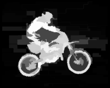}
   \end{minipage}}
   \subfigure[SF]{
   \begin{minipage}[b]{0.080\textwidth}
   \includegraphics[width=1\textwidth]{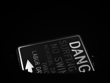}\\
   \includegraphics[width=1\textwidth]{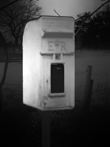}\\
   \includegraphics[width=1\textwidth]{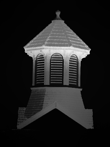}\\
   \includegraphics[width=1\textwidth]{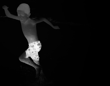}\\
   \includegraphics[width=1\textwidth]{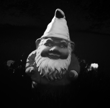}\\
   \includegraphics[width=1\textwidth]{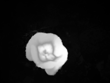}\\
   \includegraphics[width=1\textwidth]{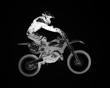}
   \end{minipage}}
   \subfigure[SVO]{
   \begin{minipage}[b]{0.080\textwidth}
  \includegraphics[width=1\textwidth]{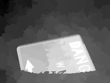}\\
   \includegraphics[width=1\textwidth]{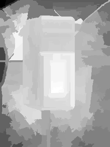}\\
   \includegraphics[width=1\textwidth]{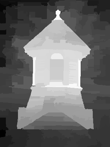}\\
   \includegraphics[width=1\textwidth]{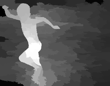}\\
   \includegraphics[width=1\textwidth]{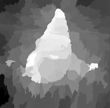}\\
   \includegraphics[width=1\textwidth]{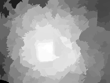}\\
   \includegraphics[width=1\textwidth]{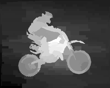}
   \end{minipage}}
   \subfigure[RBD]{
   \begin{minipage}[b]{0.080\textwidth}
   \includegraphics[width=1\textwidth]{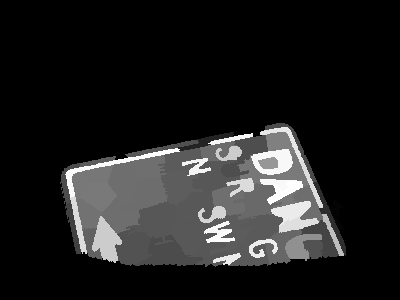}\\
   \includegraphics[width=1\textwidth]{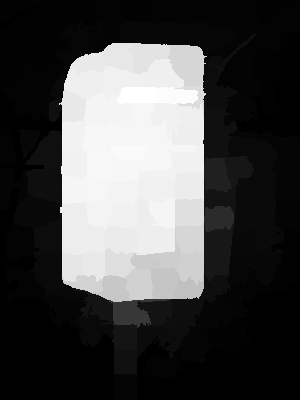}\\
   \includegraphics[width=1\textwidth]{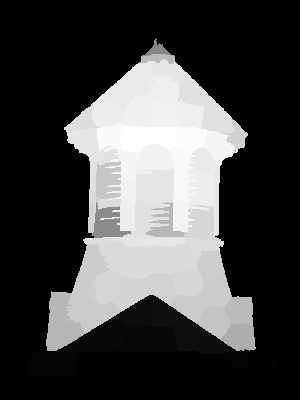}\\
   \includegraphics[width=1\textwidth]{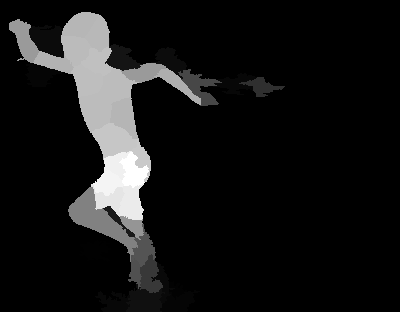}\\
   \includegraphics[width=1\textwidth]{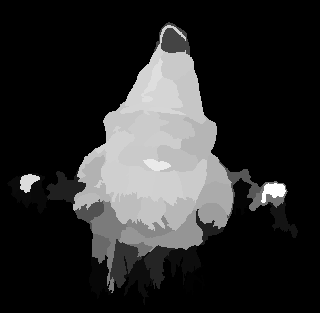}\\
   \includegraphics[width=1\textwidth]{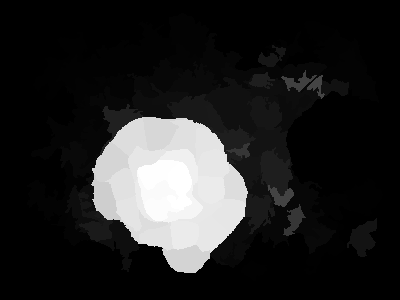}\\
   \includegraphics[width=1\textwidth]{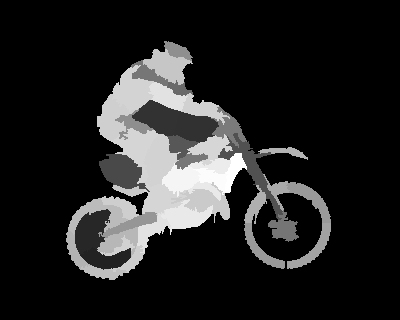}
   \end{minipage}}
   \subfigure[Ours]{
   \begin{minipage}[b]{0.080\textwidth}
   \includegraphics[width=1\textwidth]{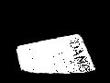}\\
   \includegraphics[width=1\textwidth]{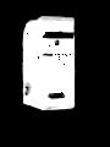}\\
   \includegraphics[width=1\textwidth]{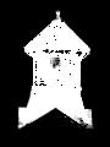}\\
   \includegraphics[width=1\textwidth]{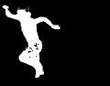}\\
   \includegraphics[width=1\textwidth]{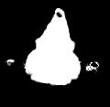}\\
   \includegraphics[width=1\textwidth]{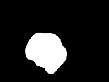}\\
   \includegraphics[width=1\textwidth]{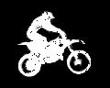}
   \end{minipage}}
  \subfigure[GT]{
   \begin{minipage}[b]{0.080\textwidth}
   \includegraphics[width=1\textwidth]{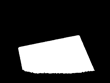}\\
   \includegraphics[width=1\textwidth]{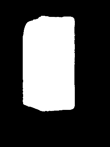}\\
   \includegraphics[width=1\textwidth]{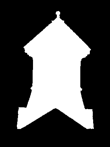}\\
   \includegraphics[width=1\textwidth]{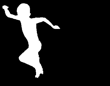}\\
   \includegraphics[width=1\textwidth]{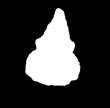}\\
   \includegraphics[width=1\textwidth]{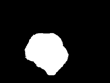}\\
   \includegraphics[width=1\textwidth]{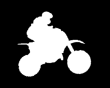}
   \end{minipage}}
\end{center}
   \vspace{-0.251cm}
   \caption{\label{ComparingSaliencyMaps}Visual comparison of saliency maps obtained by the nine competing methods (b to j) and ground truth (k).}
\end{figure*}
\begin{figure*}
\begin{center}
   \subfigure[]{\includegraphics[width=0.31\textwidth]{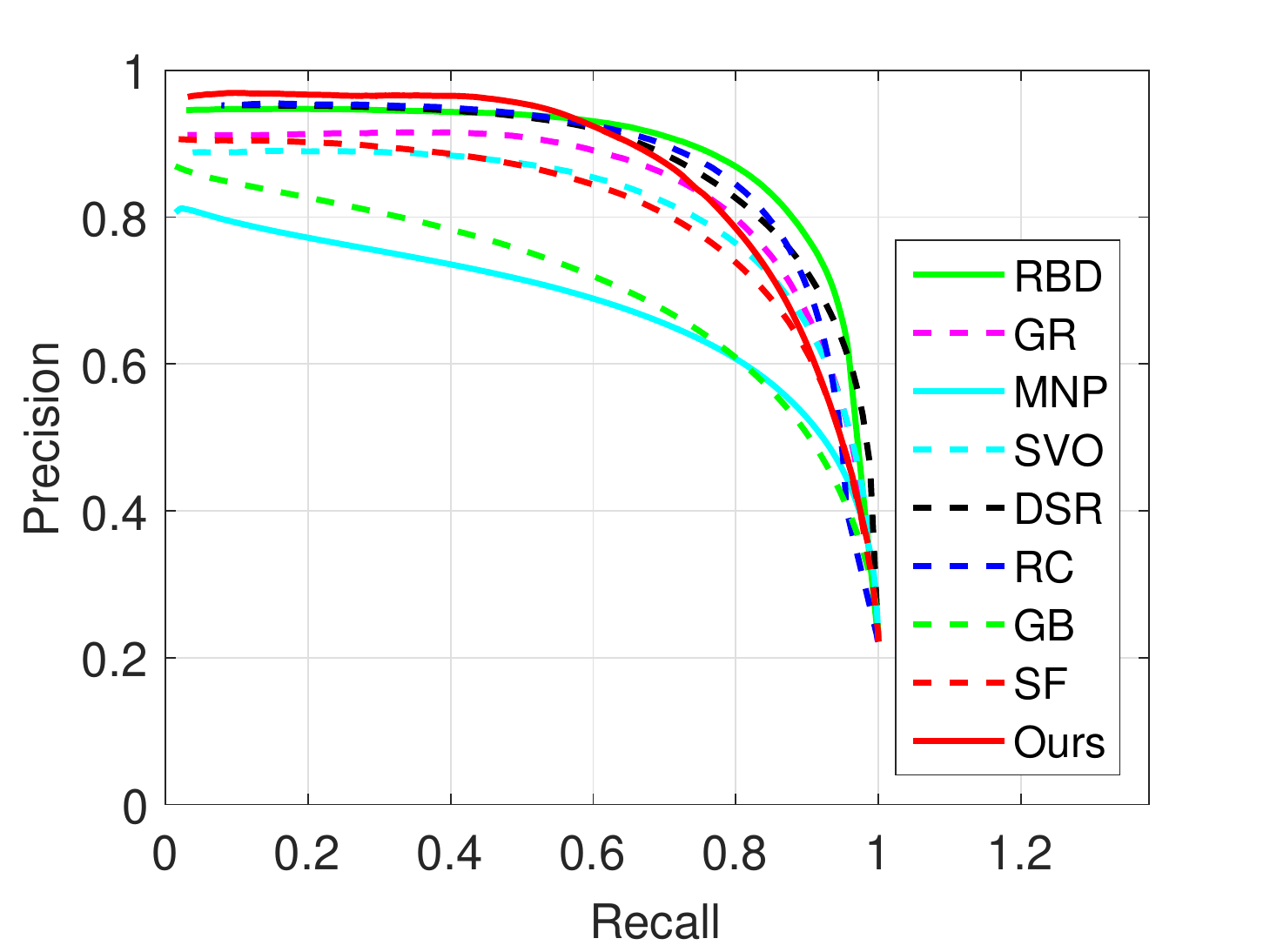}}
   \subfigure[]{\includegraphics[width=0.31\textwidth]{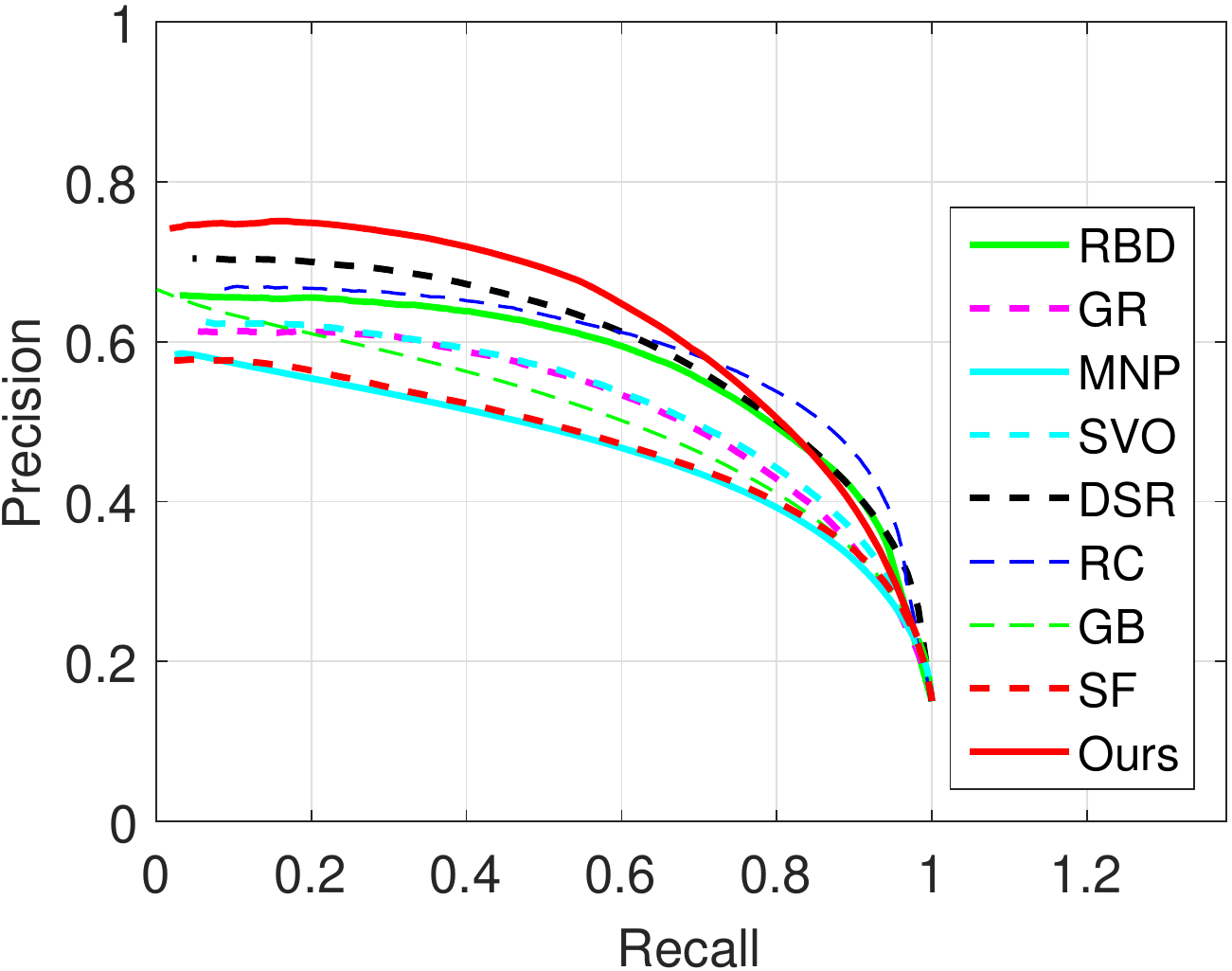}}
   \subfigure[]{\includegraphics[width=0.31\textwidth]{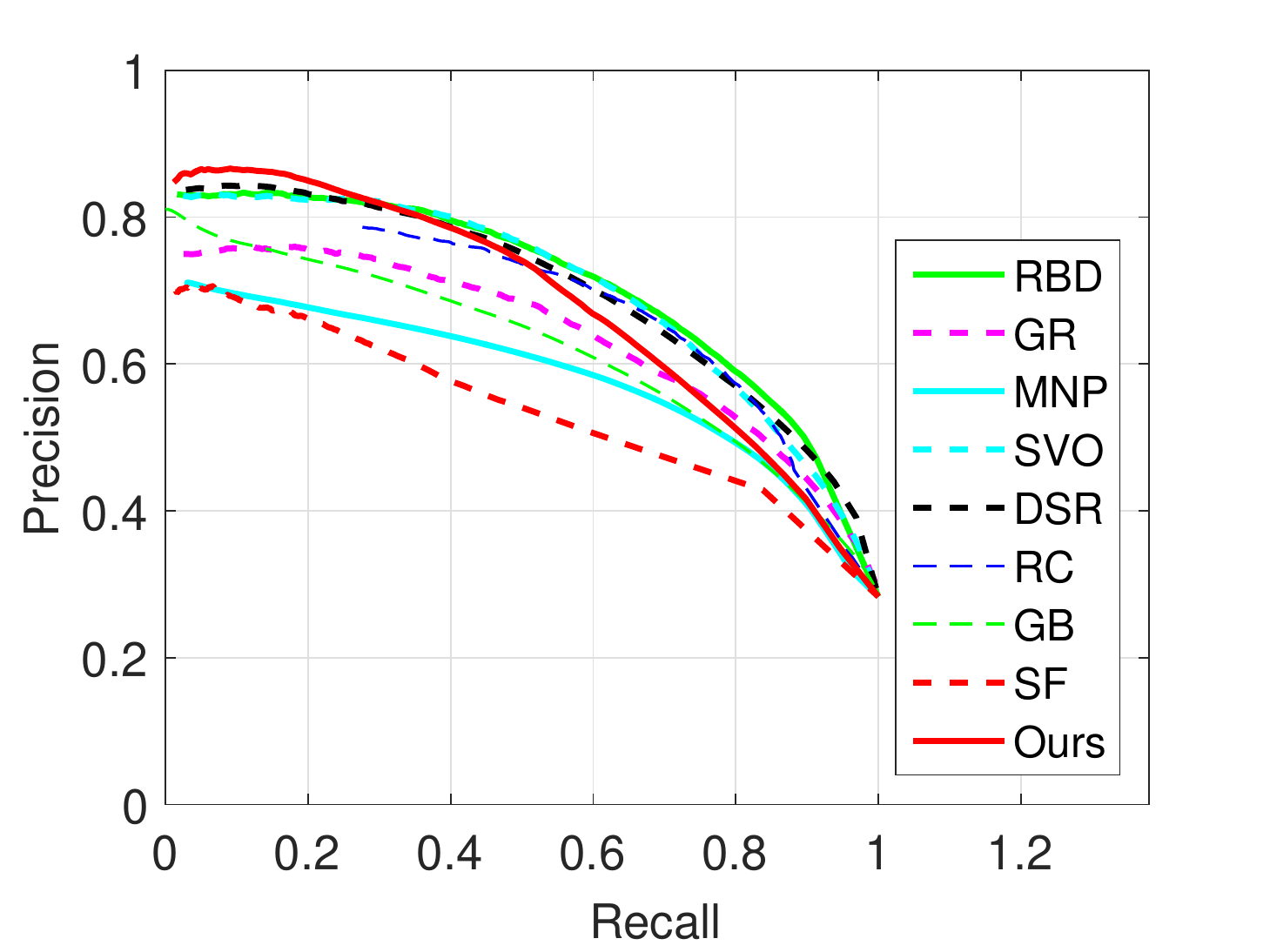}}
\end{center}
   \vspace{-0.251cm}
   \caption{\label{ComparingPRcurves} The PR curves obtained by the proposed method and the other eight saliency detection methods. (a) The results obtained by the nine competing methods on the MSRA10K dataset. (b) The results obtained by the nine competing methods on the THUK15K dataset. (c) The results obtained by the nine competing methods on the PASCAL-S dataset.}
\end{figure*}
In this section, we evaluate the performance of the proposed method on the tasks of both saliency detection and object proposal generation, and compare the proposed method with several other state-of-the-art methods.

For saliency detection, experiments are conducted on three challenging benchmark datasets: MSRA10K~\cite{LiuSalient2011}, THUR15K~\cite{ChengSalient2013} and PASCAL-S~\cite{HOU2014} (we will introduce these datasets with more details in Sec. 5.1). Precision, recall and $F_{\beta}$-measure (written as Eqn.~(\ref{FMeasure})) are adopted as the evaluation measures.
\begin{equation}\label{FMeasure}
\begin{small}
  F_\beta=\frac{(1+\beta^2)Precision \times Recall}{\beta^2 \times Precision+Recall}.
\end{small}
\end{equation}
Following~\cite{FrequencySalient2009,ChengPAMI2014}, $\beta^2$ is experimentally set to $0.3$, which emphasizes the average precision more than the average recall.

For object proposal generation, experiments are conducted on the PASCAL VOC 2007 dataset~\cite{pascal2007} and the MS COCO dataset~\cite{coco2014}. As an evaluation measure, the Intersection over Union (IoU) metric is adopted, which is written as
\begin{equation}\label{IoU}
\begin{small}
  IoU=\frac{CBB \bigcap GBB}{CBB \bigcup GBB},
\end{small}
\end{equation}
where CBB and GBB denote a candidate bounding box and a ground truth bounding box, respectively. IoU is set to 0.5, which is used as a threshold to check whether a CBB hits a GBB.

In our implementation, $\Omega_k$ is a $3{\times}3$ window. The number of superpixels used in the process of eliminating noisy regions is set to 150, which is the optimum value suggested in~\cite{saliencyFilters2012}. The constant $C$ in Eqn.~(\ref{eqn:norm}) is set to 255.  All images are resized to the fixed width (width=200px) while keeping the original aspect ratio. In the task of object proposal generation, the circumscribed rectangles of the connected and closed edges are obtained by using the build-in function \emph{regionprops} in Matlab, which measures the properties of image regions. As shown in the previous section, the proposed method obtains a set of object regions by finding the maximal cohesive clusters of image pixels. The resulting eigenvectors may not contain correct clusters if an image contains a large area of image borders. Therefore, all the borders of the test images, which have a large area of image borders, are cropped in the experiments.
\subsection{Saliency Detection}
In the first experiment, we mainly study the influence of selecting different eigenvectors on the performance of the proposed method. In order to study the contribution of different eigenvectors to the performance of the proposed method, the first six ranked eigenvectors, which correspond to the first six largest eigenvalues, are used to evaluate the performance of the proposed method. In the experiment, the value of the threshold varies from 0 to 255 to generate the precision-recall (PR) curve. Fig.~\ref{PRcurves}(a) shows that the resulting PR curves obtained by the proposed method with one of the first six ranked eigenvectors using different thresholding values. As shown in the figure, among all the six ranked eigenvectors, the contribution of the first eigenvector to saliency detection is the most significant. The performance of the proposed method drops quickly when one uses the other five eigenvectors which correspond to smaller eigenvalues. This indicates the most significant eigenvector usually corresponds to a salient object in an image since the salient foreground object commonly has a high contrast relative to the background.

In addition, we also evaluate the performance of the proposed method by combining the first several eigenvectors ranked by their corresponding eigenvalues. The obtained PR curves are shown in Fig.~\ref{PRcurves}(b). As can be seen, the performance of the proposed method with the combination of the first two eigenvectors (i.e., E1 and E2) outperforms that of the proposed method with the other combinations. Based on the analysis in Section 3.3, an eigenvector corresponding to a higher eigenvalue usually contains salient object regions with a higher probability. Thus, the proposed method using the first two ranked eigenvectors outperforms the proposed method using the other pairwise combinations of the eigenvectors. We also note that the average precision obtained by the proposed method becomes lower when it uses three or more eigenvectors. The reason is that not all the objects obtained from the first several ranked eigenvectors (especially for the lower ranked eigenvectors) are salient objects. Thus, using the first two eigenvectors (i.e., E1 and E2) for salient object detection is a good tradeoff between recall and precision.
\begin{table*}\small
\begin{center}
\caption {\label{TblMSRA10K}The average precision, average recall and $F_{\beta}$-measure obtained by the proposed method and the other eight state-of-the-arts methods for saliency detection on the MSRA10K dataset. The top three best results are highlighted by red, green and blue, respectively.}
\vspace{-0.251cm}
\begin{tabular}{|l|c|c|c|c|c|c|c|c|c|c|}
  \hline
  Methods                           & SF                 & MNP  & GB    & RBD                  & GR       & SVO             & DSR                  & RC                    & Ours\\\hline
  Average Precision                 & \color{blue}{0.86} & 0.65 & 0.69  & \color{green}{0.88}  & 0.78     & 0.59            & \color{green}{0.88}  & 0.84                  &  \color{red}{0.92}\\\hline
  Average Recall                    &0.36                &0.55  & 0.49  &\color{green}{ 0.65}   & \color{blue}{0.63}     &\color{red}{0.76}& 0.61                 &\color{green}{ 0.65}    & 0.46\\\hline
  $F_{\beta}$-measure               &0.77                &0.64  & 0.67  & \color{red}{0.86}    & 0.76     &0.60             & \color{green}{0.85}                  & \color{blue}{0.82}   & \color{green}{0.85} \\\hline
\end{tabular}
\centerline{}
\end{center}
\end{table*}
\begin{table*}\small
\begin{center}
\caption {\label{TblTHUR15K}The average precision, average recall and $F_{\beta}$-measure obtained by the proposed method and the other eight state-of-the-arts methods for saliency detection on the THUR15K dataset. The top three best results are highlighted by red, green and blue, respectively.}
\vspace{-0.251cm}
\begin{tabular}{|l|c|c|c|c|c|c|c|c|c|c|}
  \hline
  Methods                          & SF    & MNP  & GB   & RBD                 & GR        & SVO                & DSR                  & RC                           & Ours\\\hline
  Average Precision                & 0.53  &0.44  & 0.48 & \color{blue}{0.57}  & 0.50      & 0.39               & \color{green}{0.61}  & \color{blue}{0.57}         & \color{red}{0.69}\\\hline
  Average Recall                   &0.30   &\color{blue}{0.57}  & 0.53 & 0.56                & 0.55      &\color{red}{0.73}   & 0.53                 & \color{green}{0.61}        & 0.42\\\hline
  $F_{\beta}$-measure              &0.50   &0.44  & 0.48 &\color{blue}{0.57}   & 0.50      &0.41                & \color{green}{0.60}  & \color{blue}{0.57}         &\color{red}{0.65} \\\hline
\end{tabular}
\centerline{}
\end{center}
\end{table*}
\begin{table*}\small
\begin{center}
\caption {\label{TblPASCALS}The average precision, average recall and $F_{\beta}$-measure obtained by the proposed method and the other eight state-of-the-arts methods for saliency detection on the PASCAL-S dataset. The top three best results are highlighted by red, green and blue, respectively.}
\vspace{-0.251cm}
\begin{tabular}{|l|c|c|c|c|c|c|c|c|c|c|}
  \hline
  Methods                          & SF     & MNP           & GB        & RBD                  & GR                    & SVO                & DSR                      & RC                  & Ours\\\hline
  Average Precision                & 0.67   &0.62           & 0.64      & \color{blue}{0.74}   & 0.63                  & 0.73               & \color{green}{0.76}      & 0.67                & \color{red}{0.78}\\\hline
  Average Recall                   &0.14    &0.40           & 0.44      &  \color{blue}{0.47}  & \color{green}{0.49}   &\color{blue}{0.47}  & 0.42                     & \color{red}{0.51}   & 0.35\\\hline
  $F_{\beta}$-measure              &0.51    &0.59           & 0.62      &\color{blue}{0.70}    & 0.62                  &0.69              & \color{green}{0.71}      & 0.65                &\color{red}{0.72} \\\hline
\end{tabular}
\centerline{}
\end{center}
\end{table*}
Moreover, the PR curves obtained by the proposed method using the first ranked eigenvector and using the first ranked eigenvector with the proposed noisy region elimination strategy as a post-processing step are also shown in Fig.~\ref{PRcurves}(c) as a comparison. The result shows that the latter scheme achieves better average precision.

In the second experiment, the performance of the proposed method using the combined maps corresponding to the first two ranked eigenvectors and the noisy region elimination strategy is evaluated. We compare the proposed method with several other state-of-the-art saliency detection methods on three public challenging datasets: MSRA10K~\cite{ChengSalient2013}, THUR15K~\cite{ChengSalient2013} and PASCAL-S~\cite{HOU2014}. MSRA10K contains 10,000 images from the MSRA dataset~\cite{LiuSalient2011} with pixel-level saliency labeling. THUR15K contains five classes of images. In each class, there are around 3,000 images, all of which are downloaded from Flickr~\cite{flickr}. Only 6,233 binary salient masks of the corresponding images in THUR15K are given. All the 6,233 labeling salient masks and their corresponding images in the THUR15K dataset are used for evaluation. PASCAL-S contains 850 images from the PASCAL VOC 2010 dataset~\cite{pascal2007} with pixel-level saliency labeling. Comparisons are made between the proposed method and eight competing methods, namely, RBD~\cite{RBD2014}, GR~\cite{GR2013}, MNP~\cite{MNP2013}, SVO~\cite{SVO2011}, DSR~\cite{DSR2013}, RC~\cite{ChengPAMI2014}, GB~\cite{HarelGB2007}, and SF~\cite{saliencyFilters2012}. We select these methods because they are representative methods for saliency detection. The nine competing methods can be roughly divided into two categories: pixel-based methods and region-based methods. SF, MNP and GB are the pixel-based methods, while RBD, GR, RC, DSR, SVO and the proposed method belong to the region-based methods. Moreover, because some saliency detection methods for eye fixation (such as~\cite{Jian2015,han2016}) usually obtain low average precision on the MSRA10K and THUR15K datasets since they can only locate salient objects in an image instead of segmenting them, they are not used for comparison in this paper. Fig.~\ref{ComparingSaliencyMaps}(b)-(j) show the saliency maps produced by the nine competing methods for the input images shown in Fig.~\ref{ComparingSaliencyMaps}(a). Fig.~\ref{ComparingPRcurves}~(a) shows the PR curves obtained by all the nine competing methods on the MSRA10K dataset. As can be seen from Fig.~\ref{ComparingSaliencyMaps}(a), Fig.~\ref{ComparingPRcurves}(a) and Table~\ref{TblMSRA10K}, the proposed method achieves the best average precision in comparison to the other eight state-of-the-art methods. Generally, the region-based methods obtain better results than those obtained by the pixel-based methods. In the pixel-based methods, SF obtains the best average precision while it obtains the worst recall. In the region-based methods, the average precision obtained by RBD and DSR is 0.88, which is higher than that obtained by the other three region-based methods (i.e., GR, RC and SVO) but lower than that obtained by the proposed method. SVO obtains the best average recall since it fuses the cues of both objectness and visual saliency to improve the recall of salient objects. However, SVO obtains the worst average precision since the saliency maps obtained by SVO contain too many noises.

The PR curves obtained by all the nine competing methods on the THUR15K dataset are shown in Fig.~\ref{ComparingPRcurves}(b). From the figure, we can see that the proposed method outperforms most of the other methods, especially when the recall is less than 0.7. The quantitative comparison results can be seen in Table~\ref{TblTHUR15K}. In terms of average precision and average recall, the ranking of the nine competing methods on the THUR15K dataset is similar to that on the MSRA10K dataset. A difference is that the average precision obtained by SF on the THUR15K dataset is ranked worse relative to the other competing methods. The main reason is that many images in the THUR15K dataset suffer from color distortions, and SF is not robust to color distortions.

Fig.~\ref{ComparingPRcurves}(c) shows the PR curves obtained by all the nine competing methods on the PASCAL-S dataset. The quantitative comparison results can be seen in Table~\ref{TblPASCALS}. Generally, the results obtained by the nine competing methods on the PASCAL-S dataset are better than those on the THUR15K dataset, but they are worse than those on the MSRA10K dataset. The main reason is that PASCAL-S dataset contains more high-quality images than the THUR15K dataset, but it contains more challenging subjects than the MSRA10K dataset. From Table~\ref{TblPASCALS}, we can see that the average precision obtained by the proposed method is better than that obtained by all the other competing methods.

\begin{table}
\begin{center}
\caption{\label{tab:saliencyTime}The mean of the running time used by the nine competing methods on the MSRA10K dataset. The top three best results are highlighted by red, green and blue, respectively.}
\begin{tabular}{|l|c|c|}
\hline
Methods & Code & Running Time (sec)\\
\hline\hline
SF & MATLAB+C & \color{red}{0.01}\\
MNP & MATLAB+C &27.12 \\
GB & MATLAB+C &0.38 \\
RBD & MATLAB+C & \color{blue}{0.36} \\
GR & MATLAB+C &0.64\\
SVO & MATLAB+C & 57.2 \\
DSR & MATLAB+C & 4.32\\
RC & C &  \color{green}{0.14}\\
Ours & MATLAB & 3.21 \\
\hline
\end{tabular}
\end{center}
\end{table}
In practice, precision is emphasized more than recall. For this reason, the $F_\beta$ measure scores obtained by the nine methods on the MSRA10K, the THUR15K and the PASCAL-S datasets are calculated. The accurate quantitative comparison results about $F_\beta$-Measure scores obtained by the nine competing methods on the three datasets can be seen in the third row in Table~\ref{TblMSRA10K}, Table~\ref{TblTHUR15K} and Table~\ref{TblPASCALS}, respectively. From Table~\ref{TblMSRA10K}, we can see that the $F_\beta$ measure score obtained by the proposed method on the MSRA10K dataset is 0.85, which outperforms the other seven competing methods except for RBD. However, similar to SF, RBD also uses a superpixel segmentation method to segment an input image into perceptually homogeneous elements, which makes RBD not robust to color distortions. On the THUR15K dataset (shown in Table~\ref{TblTHUR15K}), the $F_\beta$-Measure score obtained by the proposed method is higher than that obtained by all the other eight competing methods. The performance of the proposed method drops on the THUR15K dataset than that on the MSRA10K dataset. This is mainly due to the lower quality of the images in the THUR15K dataset. As a consequence, the recall obtained by the proposed method drops when we only use the first two ranked eigenvectors. Compared with the other competing methods on the THUR15K dataset, similar performance drop is also observed for the performance of the other competing methods. On the PASCAL-S dataset, the rank of the nine competing methods is similar to that on the THUR15K dataset. The top three methods ranked by the $F_\beta$ measure are still the proposed method, DSR and RBD.

In terms of running time, the average execution time of the proposed method (implemented in Matlab on a 4 GHz 32GB RAM PC) for saliency detection on each image is about 3.2 seconds. Most running time is used to perform eigen-decomposition on the affinity matrix. The running time of all the nine competing methods is listed in Table~\ref{tab:saliencyTime}. As can be seen, the top three fast methods are respectively SF, RC and RBD, which are implemented in MATLAB+C or C. Although, the proposed method is implemented in Matlab only, it is more efficient than DSR, MNP and SVO. The running time of proposed method can be further reduced with more efficient implementation in C/C++.
\subsection{Object Proposal Generation}
In this section, we evaluate the performance of the proposed method for object proposal generation via the proposed cohesion measurement. The performance of the proposed method is evaluated on the PASCAL VOC 2007 dataset~\cite{pascal2007} and the MS COCO dataset~\cite{coco2014}. The PASCAL VOC 2007 dataset contains 9,963 images in which all the object instances from twenty categories are manually annotated by bounding-boxes. Since there is no training step required in the proposed method, we only report the results obtained by the proposed method on the test dataset of the PASCAL VOC 2007 dataset, which consists of 2,507 images. The MS COCO dataset consists of 91 common object categories and 40,137 images in total. All the 40,137 images are used in the experiments.

The IoU metric is used as the evaluation criterion for object proposal generation.
In the implementation, we use the first eighty ranked eigenvectors and all the pairwise combinations of the first six ranked eigenvectors to generate object proposals. If the IoU between two candidate bounding boxes (CBBs) is greater than 0.95, the CBB with a lower objectness score is removed from the pairwise CBBs. Then each of the remaining CBBs is assigned with an objectness score using the truncated objectness measure. All the CBBs, whose objectness scores are zero, are also removed. The resulting ranked CBBs are used for evaluation.

\begin{figure}
\begin{center}
   \includegraphics[width=0.80\linewidth]{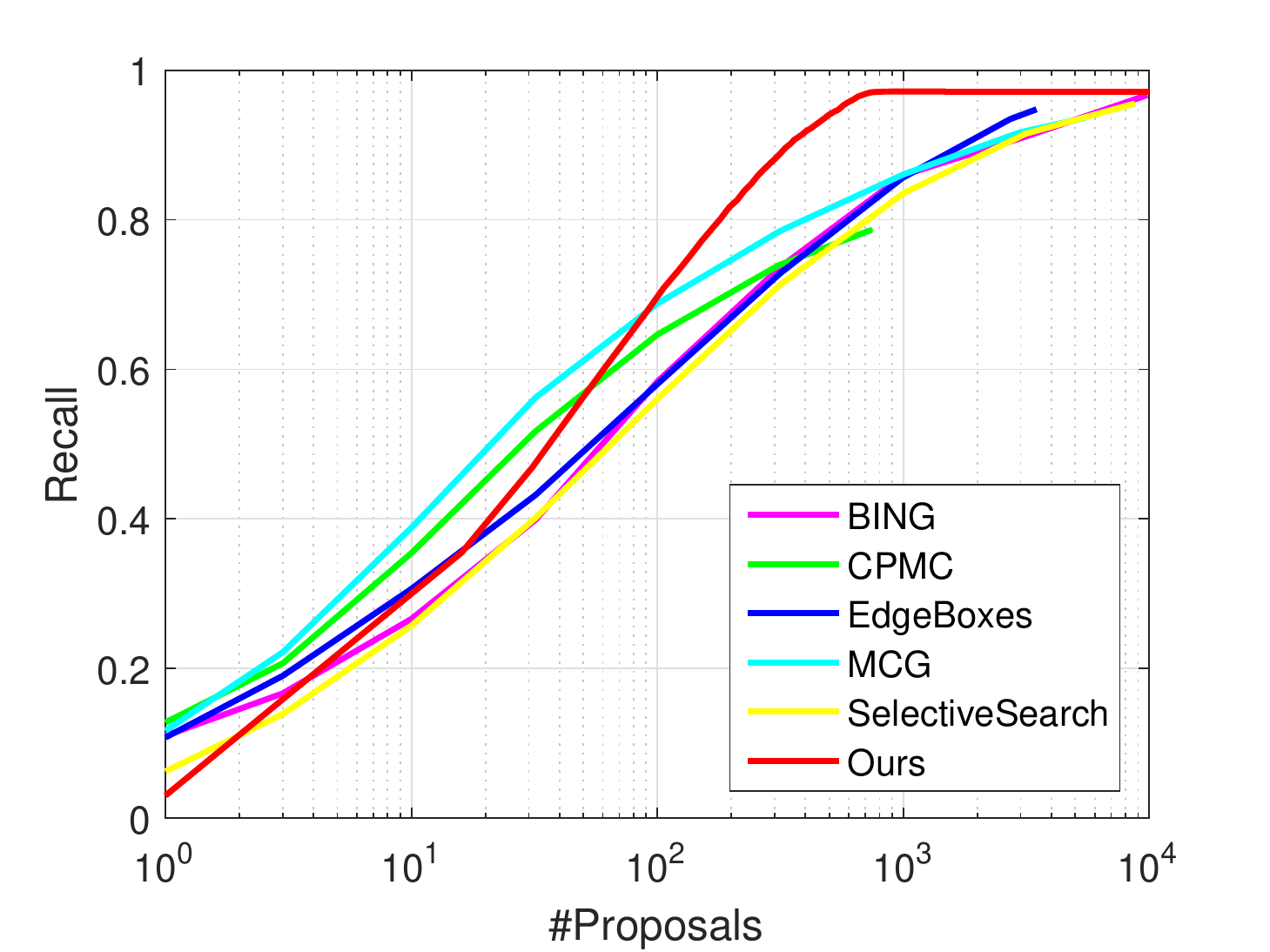}
\end{center}
   \vspace{-0.251cm}
   \caption{The trade-off between the number of object proposals in each image and the recall for object proposal generation on the Pascal VOC 2007 dataset. Selective Search~\cite{SelectiveSearch13}, CPMC~\cite{CPMC2012}, BING~\cite{BingObj2014}, EdgeBoxes~\cite{edgeBoxes2014} and MCG~\cite{MCG2014} are compared with the proposed method.}
\label{Fig:ProposalCurves}
\end{figure}
\begin{figure}
\begin{center}
   \includegraphics[width=0.80\linewidth]{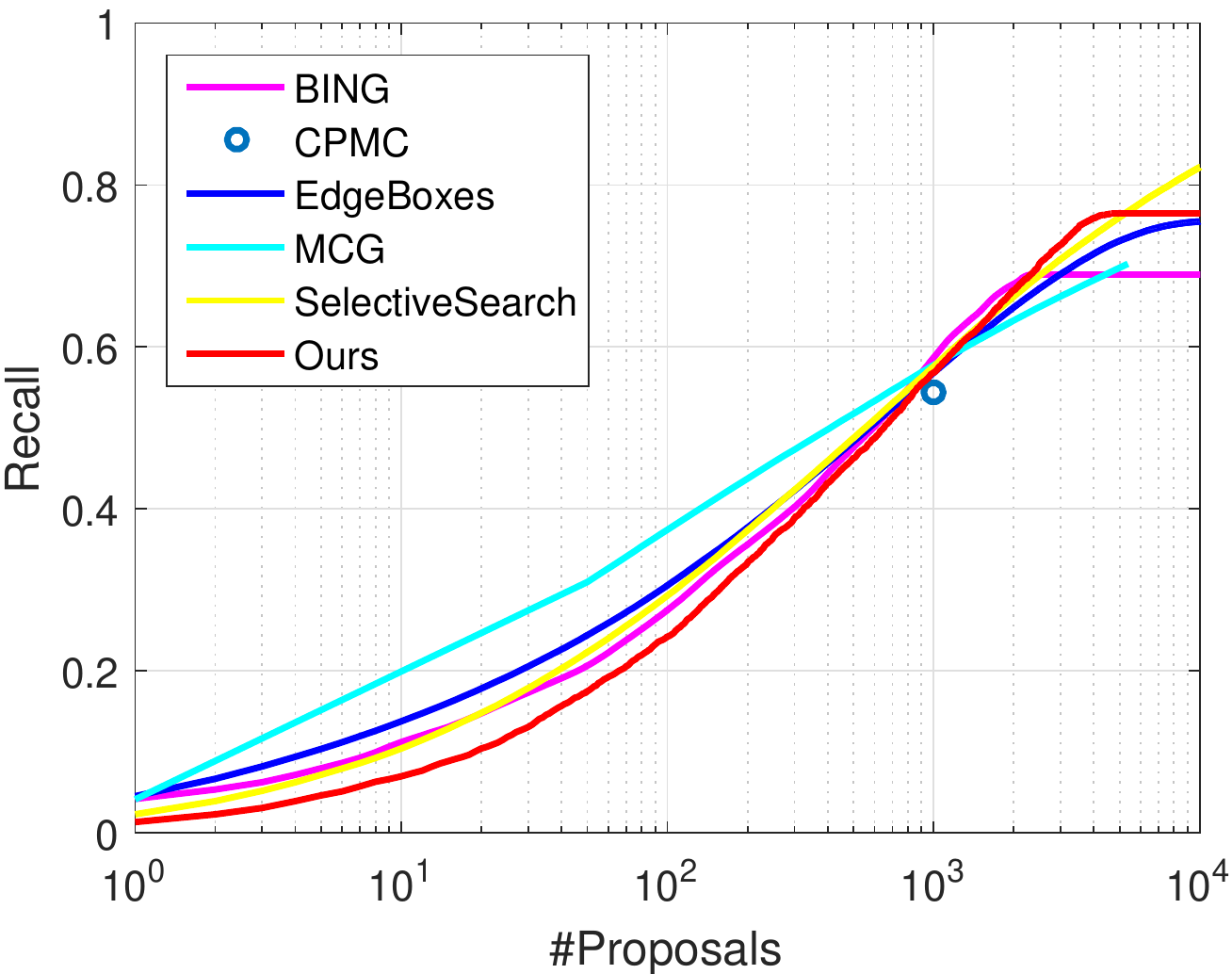}
\end{center}
   \vspace{-0.251cm}
   \caption{The trade-off between the number of object proposals in each image and the recall for object proposal generation on the MS COCO dataset. Selective Search~\cite{SelectiveSearch13}, CPMC~\cite{CPMC2012}, BING~\cite{BingObj2014}, EdgeBoxes~\cite{edgeBoxes2014} and MCG~\cite{MCG2014} are compared with the proposed method.}
\label{Fig:cocoProposalCurves}
\end{figure}
We compare the proposed method with several state-of-the-art methods for object proposal generation, and evaluate the recall obtained by the competing methods when varying the number of  object proposals. We select the following five methods for comparison: Selective Search~\cite{SelectiveSearch13}, CPMC~\cite{CPMC2012}, BING~\cite{BingObj2014}, EdgeBoxes~\cite{edgeBoxes2014} and MCG~\cite{MCG2014}. The five competing methods are selected since they are related to the proposed method: Selective Search, CPMC and MCG are based on image segmentation; BING and EdgeBoxes are based on the boundary or edge feature. Fig.~\ref{Fig:ProposalCurves} shows the comparison results on the PASCAL VOC 2007 dataset.  The recall obtained by the proposed method is not high when using only a small number of proposals, which is caused by the following two reasons. The first reason is that the proposed truncated objectness measure is only used to eliminate noisy regions rather than to measure the objectness of an object. Thus, the top dozens of object proposals ranked by the proposed truncated objectness measure contain a small number of objects. The second reason is that a small object in an image usually obtains a small objectness value by using the proposed truncated objectness measure since the small object is prone to be treated as a noisy region. However, the proposed method achieves the best recall when the number of the proposals is more than 600.

\begin{figure*}
\begin{center}
   \includegraphics[width=0.98\linewidth]{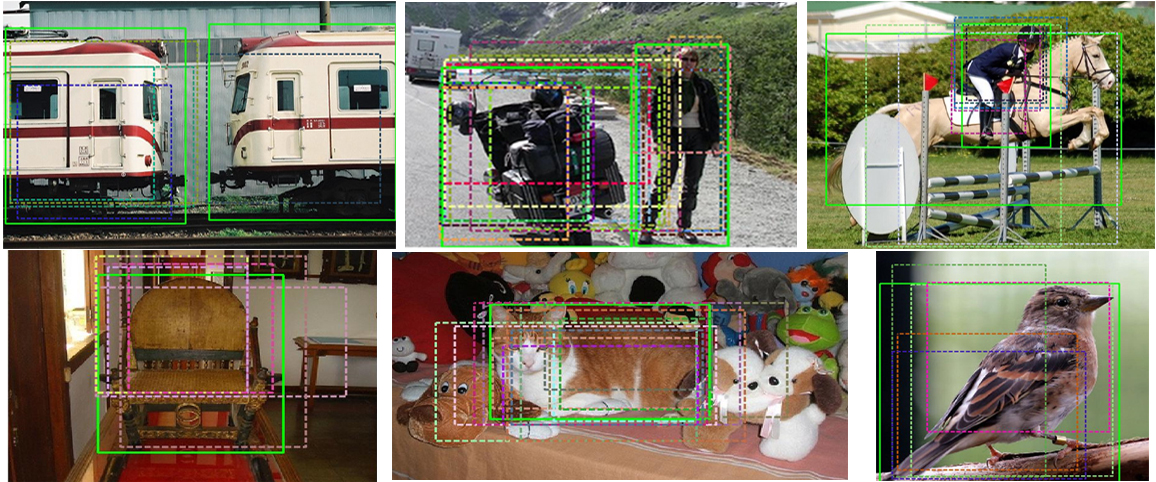}
\end{center}
   \vspace{-0.251cm}
   \caption{Qualitative examples showing the results obtained by of the proposed method. The GBBs are shown in the green solid lines and the correct CBBs with IoU higher than 0.5 are shown in the dash lines with different colors.}
\label{Fig:ProposalRe}
\end{figure*}

\begin{table}
\begin{center}
\caption{\label{tab:conf}The average running time of the six competing methods on the Pascal VOC 2007 object detection dataset for object proposal generation. The top three best results are highlighted by red, green and blue, respectively.}
\begin{tabular}{|l|c|c|}
\hline
Methods & Code & Running Time (sec)\\
\hline\hline
BING & C & \color{red}{0.003}\\
CPMC & MATLAB+C & $>$200 \\
EdgeBoxes & MATLAB+C &\color{green}{0.35} \\
MCG & MATLAB+C & 31.13 \\
SelectiveSearch & MATLAB+C &13.25\\
Ours & MATLAB & \color{blue}{12.57} \\
\hline
\end{tabular}
\end{center}
\end{table}
Fig.~\ref{Fig:cocoProposalCurves} shows the comparison results on the MS COCO dataset. As can be seen, the proposed method obtains higher recall than the other competing methods except for Selective Search when the number of the proposals is more than 4,000. Many images in the MS COCO dataset contain various styles of borders which significantly affect the performance of the proposed method. Some qualitative results obtained by the proposed method for object proposal generation are shown in Fig.~\ref{Fig:ProposalRe}. Compared with the other state-of-the-art object proposal generation methods, no training step is required in the proposed method. Only two primitive features (i.e., color and edge) are used to locate the objects in an image. Since the proposed method fuses the two objectness cues, it obtains better recall than that obtained by most of the other most competing methods when using a large number of proposals.
\begin{figure*}
\begin{center}
   \subfigure[]{
   \begin{minipage}[b]{0.3\textwidth}
   \includegraphics[width=1\textwidth]{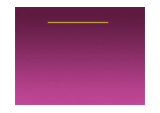}
   \end{minipage}}
   \subfigure[]{
   \begin{minipage}[b]{0.3\textwidth}
   \includegraphics[width=1\textwidth]{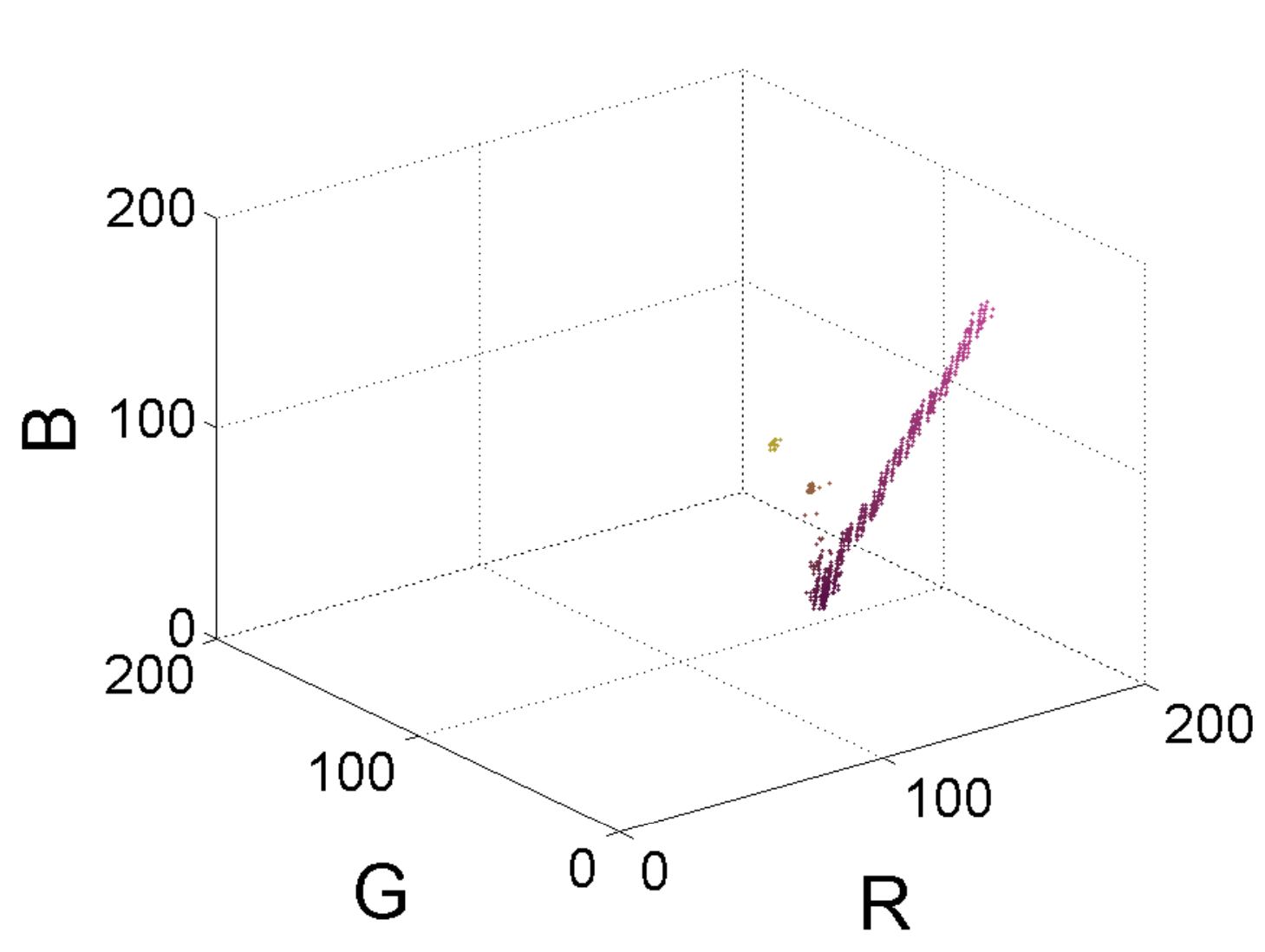}
   \end{minipage}}
   \subfigure[]{
   \begin{minipage}[b]{0.3\textwidth}
   \includegraphics[width=1\textwidth]{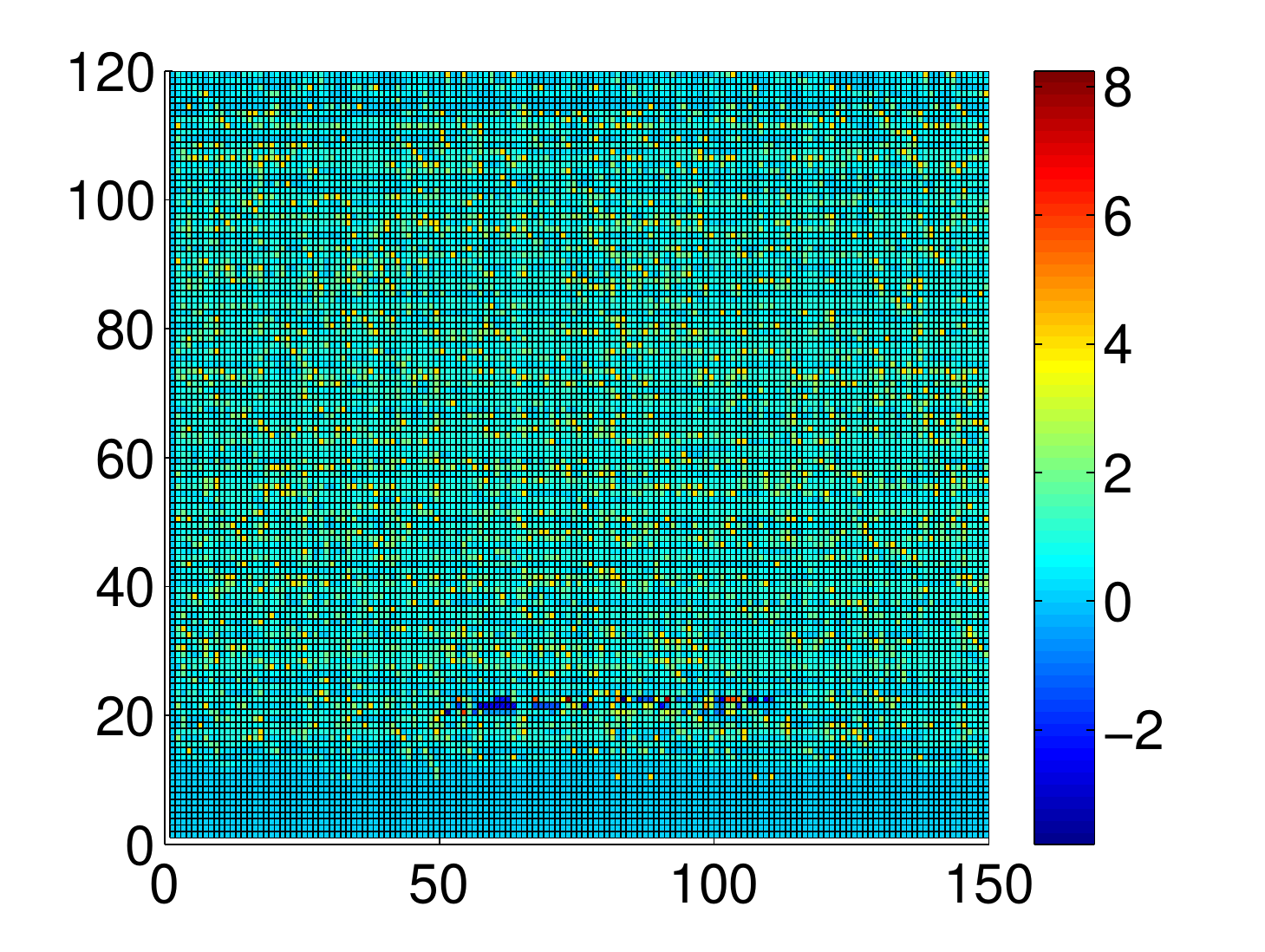}
   \end{minipage}}
\end{center}
   \vspace{-0.251cm}
   \caption{\label{Fig::affinityEvidence}The visualization of the affinity values obtained by using Eqn.~(\ref{innerProduct}). (a) The input image. (b) The color line of the input image. (c) The visualization of the affinity map obtained by using Eqn.~(\ref{innerProduct}). More explanations can be found in \textbf{Appendix B}.}
\end{figure*}

The running time of the competing object proposal generation methods is listed in Table~\ref{tab:conf}. The top three fastest methods are BING, EdgeBoxes and the proposed method, respectively. However, the implementation of the proposed method is totally in Matlab without interfacing with C as BING and  EdgeBoxes do. Generally, the running time of the edge based methods is lower than that of the image segmentation based methods. The slowest method is CPMC, which requires more than 200 seconds to deal with each image in the Pascal VOC 2007 dataset on average.

\section{Conclusion}
In this paper, we have proposed a novel method to discover objects latent in an image. We propose to encode cohesion measurement into a new affinity matrix, which is robust to color distortions. Images pixels from the same object region are more effectively clustered by using the formulated affinity matrix compared with the traditional spectral clustering methods. Moreover, the problem of object discovery is solved by utilizing the eigenvectors of the proposed affinity matrix, and the corresponding eigenvalues turn out to be an appropriate cohesion measurement for detecting a object region. In addition, we have also shown that how the proposed method is applied to the tasks of saliency detection and object proposal generation. As shown in the experiments, the proposed method is more effective than most of the other state-of-the-art methods on the two tasks. Moreover, since the proposed method dose not require a training step, it can be directly applied to the other related computer vision tasks, such as image segmentation, object tracking, etc.\\


%

\textbf{Acknowledgement:} This work was supported by the National Natural Science Foundation of China under Grants U1605252, 61472334, 61572408 and 61571379, and by the Key Research Program of the Chinese Academy of Sciences under Grant KGZD-EW-T03.
\vfill
\appendices
\appendices
\numberwithin{equation}{section}
\section{Derivation of Eqn.~(\ref{innerProduct})}
Since the covariance matrix $\Sigma_k$ is a symmetric and positive semidefinite matrix, the covariance matrix $\Sigma_k$ can be decomposed by using the SVD decomposition method as follow:
\begin{equation}\label{eq::SVD}
  \Sigma_k=U\Lambda U^T,
\end{equation}
  where $U$ is an orthogonal matrix whose each column is the eigenvector of the covariance matrix $\Sigma_k$ and $\Lambda$ is a diagonal matrix with non-negative diagonal entries $\phi_z$. $\phi_z$ is the $z$th eigenvalue of the covariance matrix $\Sigma_k$. Then
  \begin{equation}\label{eq::deriveSVD}
  \begin{split}
  (\Sigma_k+\tau E)^{-1}=(U\Lambda U^T+\tau E)^{-1} \\
  =(U(\Lambda+\tau E) U^T)^{-1},
  \end{split}
  \end{equation}
  where $E$ is an identity matrix. Using the fact that $U^{-1}=U^T$, we have,
  \begin{equation}\label{eq::deriveSVD2}
    (U(\Lambda+\tau E) U^T)^{-1}=U^T(\Lambda+\tau E)^{-1} U.
  \end{equation}
  Let $\tilde{p_i}=(p_i-\mu_k)$ and $\tilde{p_j}=(p_j-\mu_k)$ denote the new representations of pixels $p_i$ and $p_j$, respectively. Then, $A_E$ can be transformed into
  \begin{equation}\label{eq::SVDFinal}
  \begin{split}
   A_E=\tilde{p_i}^T U^T(\Lambda+\tau E)^{-1}U\tilde{p_i}\\
   =(U\tilde{p_i})^T (\Lambda+\tau E)^{-1} (U\tilde{p_i}) \\
   =\theta \cdot (U\tilde{p_i})^T (U\tilde{p_i})\\
   =\theta \cdot (U(p_i-\mu_k))^T (U(p_j-\mu_k)),
   \end{split}
   \end{equation}
   where $0<\theta_z=\frac{1}{\phi_z+\tau}$.

\section{An Illustration of the Affinity Values Obtained by Using Eqn.~(\ref{innerProduct})}
To show that $A_E$ is robust to variations in lighting and shadows, we give an illustration of the affinity map obtained by using $A_E$. As shown in Fig.~\ref{Fig::affinityEvidence}(a), the synthetic rectangle contains two real-world colors: magenta and orange. The magenta color undergoes intensity variation while the orange color keeps constant. The color line of the rectangle is shown in Fig.~\ref{Fig::affinityEvidence}(b). As can be seen, the magenta pixels distribute in an elongated ellipse region in the RGB color space. We calculate the affinity value between each pair of pixels in a $3{\times}3$ window by using $A_E$ (i.e., Eqn.~(\ref{innerProduct})). For easy visualization, we only record the affinity value between the fixed pair of pixels in each small window. For the $k$th (remind that $k= x \cdot w + y$) window, the affinity value between the pixel at $(x-1,y-1)$ and the pixel at $(x+1,y+1)$ is recorded. The affinity value at $(x-1,y-1)$ in the affinity map is thus obtained. The resulting affinity map is shown in Fig.~\ref{Fig::affinityEvidence}(c). As can be seen, although the magenta color undergoes intensity variation, the obtained affinity values of the magenta pixels are almost identical and they are all greater than zero, which imply the magenta pixels inside the rectangle region are similar. However, most of the obtained affinity values between the magenta pixels and the orange pixels are smaller than zero, which indicates that the two colors are two different real-world colors.

\ifCLASSOPTIONcompsoc
\else
\fi

\ifCLASSOPTIONcaptionsoff
  \newpage
\fi



%

%
%

\bibliographystyle{IEEEtran}
\bibliography{PAMIegbib}

\begin{thebibliography}{10}
\providecommand{\url}[1]{#1}
\csname url@samestyle\endcsname
\providecommand{\newblock}{\relax}
\providecommand{\bibinfo}[2]{#2}
\providecommand{\BIBentrySTDinterwordspacing}{\spaceskip=0pt\relax}
\providecommand{\BIBentryALTinterwordstretchfactor}{4}
\providecommand{\BIBentryALTinterwordspacing}{\spaceskip=\fontdimen2\font plus
\BIBentryALTinterwordstretchfactor\fontdimen3\font minus
  \fontdimen4\font\relax}
\providecommand{\BIBforeignlanguage}[2]{{%
\expandafter\ifx\csname l@#1\endcsname\relax
\typeout{** WARNING: IEEEtran.bst: No hyphenation pattern has been}%
\typeout{** loaded for the language `#1'. Using the pattern for}%
\typeout{** the default language instead.}%
\else
\language=\csname l@#1\endcsname
\fi
#2}}
\providecommand{\BIBdecl}{\relax}
\BIBdecl

\bibitem{Viola2001}
P.~Viola and M.~Jones, ``Robust real-time object detection,'' in \emph{Int. J.
  Comput. Vis. (IJCV)}, vol.~57, no.~2, 2001, pp. 137--154.

\bibitem{HOG2005}
N.~Dalal and B.~Triggs, ``Histograms of oriented gradients for human
  detection,'' in \emph{Proc. IEEE Conf. Comput. Vis. Pattern Recognit.
  (CVPR)}, 2005, pp. 886--893.

\bibitem{YangC14}
Y.~Pang, K.~Zhang, Y.~Yuan, and K.~Wang, ``Distributed object detection with
  linear svms,'' \emph{{IEEE} Trans. Cybernetics (TYCB)}, vol.~44, no.~11, pp.
  2122--2133, 2014.

\bibitem{SelectiveSearch13}
J.~Uijlings, K.~van~de Sande, T.~Gevers, and A.~Smeulders, ``Selective search
  for object recognition,'' \emph{Int. J. Comput. Vis. (IJCV)}, vol. 104,
  no.~2, pp. 154--171, 2013.

\bibitem{xuelong2017}
Y.~Pang, J.~Cao, and X.~Li, ``Learning sampling distributions for efficient
  object detection,'' \emph{IEEE Transactions on Cybernetics (TYCB)}, vol.~47,
  no.~1, pp. 117--129, 2017.

\bibitem{Rcnn2014}
R.~Girshick, J.~Donahue, T.~Darrell, and J.~Malik, ``Rich feature hierarchies
  for accurate object detection and semantic segmentation,'' in \emph{Proc.
  IEEE Conf. Comput. Vis. Pattern Recognit. (CVPR)}, 2014, pp. 580--587.

\bibitem{Sppnet2014}
K.~He, X.~Zhang, S.~Ren, and J.~Sun, ``Spatial pyramid pooling in deep
  convolutional networks for visual recognition,'' in \emph{Proc. Eur. Comput.
  Vis. Conf. (ECCV)}, 2014.

\bibitem{Objectness2012}
B.~Alexe, T.~Deselaers, and V.~Ferrari, ``Measuring the objectness of image
  windows,'' \emph{IEEE Trans. Pattern Anal. Mach. Intell. (TPAMI)}, vol.~34,
  no.~11, pp. 2189--2202, 2012.

\bibitem{BingObj2014}
M.-M. Cheng, Z.~Zhang, W.-Y. Lin, and P.~H.~S. Torr, ``{BING}: Binarized normed
  gradients for objectness estimation at 300fps,'' in \emph{Proc. IEEE Conf.
  Comput. Vis. Pattern Recognit. (CVPR)}, 2014, pp. 3286--3293.

\bibitem{FangTIP12}
Y.~Fang, Z.~Chen, W.~Lin, and C.-W. Lin, ``Saliency detection in the compressed
  domain for adaptive image retargeting,'' \emph{IEEE Trans. Image Processing
  (TIP)}, vol.~21, no.~9, pp. 3888--3901, 2012.

\bibitem{fang2014}
Y.~Fang, W.~Lin, Z.~Chen, C.~M. Tsai, and C.~W. Lin, ``A video saliency
  detection model in compressed domain,'' \emph{IEEE Transactions on Circuits
  and Systems for Video Technology (TCSVT)}, vol.~24, no.~1, pp. 27--38, 2014.

\bibitem{cao2014}
X.~Cao, Z.~Tao, B.~Zhang, H.~Fu, and W.~Feng, ``Self-adaptively weighted
  co-saliency detection via rank constraint,'' \emph{{IEEE} Trans. Image
  Processing (TIP)}, vol.~23, no.~9, pp. 4175--4186, 2014.

\bibitem{VisualAttention2013}
A.~Borji and L.~Itti, ``State-of-the-art in visual attention modeling,''
  \emph{IEEE Trans. Pattern Anal. Mach. Intell. (TPAMI)}, vol.~35, no.~1, pp.
  185--207, 2013.

\bibitem{huazhu2013}
H.~Fu, X.~Cao, and Z.~Tu, ``Cluster-based co-saliency detection,'' \emph{{IEEE}
  Trans. Image Processing (TIP)}, vol.~22, no.~10, pp. 3766--3778, 2013.

\bibitem{ColorLine2004}
I.~Omer and M.~Werman, ``Color lines: Image specific color representation,'' in
  \emph{Proc. IEEE Conf. Comput. Vis. Pattern Recognit. (CVPR)}, 2004, pp.
  946--953.

\bibitem{Shi2000}
J.~Shi and J.~Malik, ``Normalized cuts and image segmentation,'' \emph{IEEE
  Trans. Pattern Anal. Mach. Intell. (TPAMI)}, vol.~22, no.~8, pp. 888--905,
  2000.

\bibitem{Freeman1998}
P.~Perona and W.~Freeman, ``A factorization approach to grouping,'' in
  \emph{Proc. Eur. Comput. Vis. Conf. (ECCV)}, 1998, pp. 655--670.

\bibitem{Ng01}
A.~Y. Ng, M.~I. Jordan, and Y.~Weiss, ``On spectral clustering: Analysis and an
  algorithm,'' in \emph{Proc. Adv. Neural Inf. Process. Syst. (NIPS)}.\hskip
  1em plus 0.5em minus 0.4em\relax MIT Press, 2001, pp. 849--856.

\bibitem{biggs1993}
N.~Biggs, \emph{Algebraic Graph Theory}, 2nd~ed.\hskip 1em plus 0.5em minus
  0.4em\relax Cambridge University Press, 1993.

\bibitem{LPP2003}
X.~He and P.~Niyogi, ``Locality preserving projections,'' in \emph{Proc. Adv.
  Neural Inf. Process. Syst. (NIPS)}.\hskip 1em plus 0.5em minus 0.4em\relax
  MIT Press, 2003.

\bibitem{CaiC15}
D.~Cai and X.~Chen, ``Large scale spectral clustering via landmark-based sparse
  representation,'' \emph{{IEEE} Trans. Cybernetics (TYCB)}, vol.~45, no.~8,
  pp. 1669--1680, 2015.

\bibitem{YangC15}
Y.~Yang, Z.~Ma, Y.~Yang, F.~Nie, and H.~T. Shen, ``Multitask spectral
  clustering by exploring intertask correlation,'' \emph{{IEEE} Trans.
  Cybernetics (TYCB)}, vol.~45, no.~5, pp. 1069--1080, 2015.

\bibitem{Levin2006}
A.~Levin, D.~Lischinski, and Y.~Weiss, ``A closed form solution to natural
  image matting,'' in \emph{Proc. IEEE Conf. Comput. Vis. Pattern Recognit.
  (CVPR)}, 2006, pp. 61--68.

\bibitem{Gonzalez2006}
R.~C. Gonzalez and R.~E. Woods, \emph{Digital Image Processing (3rd
  Edition)}.\hskip 1em plus 0.5em minus 0.4em\relax Prentice-Hall, Inc., 2006.

\bibitem{Cristianini02}
N.~Cristianini, J.~Kandola, A.~Elisseeff, and J.~Shawe-Taylor, ``On
  kernel-target alignment,'' in \emph{Proc. Adv. Neural Inf. Process. Syst.
  (NIPS)}.\hskip 1em plus 0.5em minus 0.4em\relax MIT Press, 2002, pp.
  367--373.

\bibitem{Horn1986}
R.~A. Horn and C.~R. Johnson, Eds., \emph{Matrix Analysis}.\hskip 1em plus
  0.5em minus 0.4em\relax New York, NY, USA: Cambridge University Press, 1986.

\bibitem{saliencyFilters2012}
F.~Perazzi, Y.~Pritch, and A.~Hornung, ``Saliency filters: Contrast based
  filtering for salient region detection,'' in \emph{Proc. IEEE Conf. Comput.
  Vis. Pattern Recognit. (CVPR)}, 2012, pp. 733--740.

\bibitem{FrequencySalient2009}
R.~Achanta, S.~S. Hemami, F.~J. Estrada, and S.~S¨¹sstrunk, ``Frequency-tuned
  salient region detection,'' in \emph{Proc. IEEE Conf. Comput. Vis. Pattern
  Recognit. (CVPR)}, 2009, pp. 1597--1604.

\bibitem{ChengPAMI2014}
M.-M. Cheng, N.~J. Mitra, X.~Huang, P.~H.~S. Torr, and S.-M. Hu, ``Global
  contrast based salient region detection,'' \emph{IEEE Trans. Pattern Anal.
  Mach. Intell. (TPAMI)}, vol.~37, no.~3, pp. 569--582, 2014.

\bibitem{SLIC2012}
R.~Achanta, A.~Shaji, K.~Smith, A.~Lucchi, P.~Fua, and S.~Susstrunk, ``Slic
  superpixels compared to state-of-the-art superpixel methods,'' \emph{IEEE
  Trans. Pattern Anal. Mach. Intell. (TPAMI)}, vol.~34, no.~11, pp. 2274--2282,
  2012.

\bibitem{edgeBoxes2014}
C.~L. Zitnick and P.~Doll\'{a}r, ``{Edge Boxes: Locating Object Proposals from
  Edges},'' in \emph{Proc. Eur. Comput. Vis. Conf. (ECCV)}, 2014, pp. 391--405.

\bibitem{gPbowtucm2011}
P.~Arbelaez, M.~Maire, C.~Fowlkes, and J.~Malik, ``Contour detection and
  hierarchical image segmentation,'' \emph{IEEE Trans. Pattern Anal. Mach.
  Intell. (TPAMI)}, vol.~33, no.~5, pp. 898--916, 2011.

\bibitem{cannyEdge1986}
J.~Canny, ``A computational approach to edge detection,'' \emph{IEEE Trans.
  Pattern Anal. Mach. Intell. (TPAMI)}, vol.~8, no.~6, pp. 679--698, 1986.

\bibitem{LiuSalient2011}
T.~Liu, Z.~Yuan, J.~Sun, J.~Wang, N.~Zheng, X.~Tang, and H.-Y. Shum, ``Learning
  to detect a salient object,'' \emph{IEEE Trans. Pattern Anal. Mach. Intell.
  (TPAMI)}, vol.~33, no.~2, pp. 353--367, 2011.

\bibitem{ChengSalient2013}
M.-M. Cheng, N.~J. Mitra, X.~Huang, and S.-M. Hu, ``Salientshape: Group
  saliency in image collections,'' \emph{The Visual Computer}, vol.~30, no.~4,
  pp. 443--453, 2013.

\bibitem{HOU2014}
Y.~Li, X.~Hou, C.~Koch, J.~M. Rehg, and A.~L. Yuille, ``The secrets of salient
  object segmentation,'' in \emph{Proc. IEEE Conf. Comput. Vis. Pattern
  Recognit. (CVPR)}, 2014, pp. 280--287.

\bibitem{pascal2007}
M.~Everingham, L.~Gool, C.~K. Williams, J.~Winn, and A.~Zisserman, ``The pascal
  visual object classes (voc) challenge,'' \emph{Int. J. Comput. Vis. (IJCV)},
  vol.~88, no.~2, pp. 303--338, 2010.

\bibitem{coco2014}
T.~Lin, M.~Maire, S.~J. Belongie, L.~D. Bourdev, R.~B. Girshick, J.~Hays,
  P.~Perona, D.~Ramanan, P.~Doll{\'{a}}r, and C.~L. Zitnick, ``Microsoft
  {COCO:} common objects in context,'' \emph{CoRR}, vol. abs/1405.0312, 2014.

\bibitem{flickr}
``Flickr,'' \url{https://www.flickr.com/}.

\bibitem{RBD2014}
W.~Zhu, S.~Liang, Y.~Wei, and J.~Sun, ``Saliency optimization from robust
  background detection,'' in \emph{Proc. IEEE Conf. Comput. Vis. Pattern
  Recognit. (CVPR)}, 2014, pp. 2814--2821.

\bibitem{GR2013}
C.~Yang, L.~Zhang, and H.~Lu, ``Graph-regularized saliency detection with
  convex-hull-based center prior,'' \emph{IEEE Signal Processing Letters},
  vol.~20, pp. 637--640, 2013.

\bibitem{MNP2013}
R.~Margolin, L.~Zelnik-Manor, and A.~Tal, ``Saliency for image manipulation,''
  \emph{The Visual Computer}, pp. 1--12, 2013.

\bibitem{SVO2011}
K.-Y. Chang, T.-L. Liu, H.-T. Chen, and S.-H. Lai, ``Fusing generic objectness
  and visual saliency for salient object detection,'' in \emph{Proc. IEEE Int.
  Conf. Comput. Vis. (ICCV)}, 2011.

\bibitem{DSR2013}
X.~Li, H.~Lu, L.~Zhang, X.~Ruan, and M.-H. Yang, ``Saliency detection via dense
  and sparse reconstruction,'' in \emph{Proc. IEEE Int. Conf. Comput. Vis.
  (ICCV)}, 2013, pp. 2976--2983.

\bibitem{HarelGB2007}
J.~Harel, C.~Koch, and P.~Perona, ``Graph-based visual saliency,'' in
  \emph{Proc. Adv. Neural Inf. Process. Syst. (NIPS)}, 2007, pp. 545--552.

\bibitem{Jian2015}
M.~Jian, K.~Lam, J.~Dong, and L.~Shen, ``Visual-patch-attention-aware saliency
  detection,'' \emph{{IEEE} Trans. Cybernetics (TYCB)}, vol.~45, no.~8, pp.
  1575--1586, 2015.

\bibitem{han2016}
J.~Han, D.~Zhang, S.~Wen, L.~Guo, T.~Liu, and X.~Li, ``Two-stage learning to
  predict human eye fixations via sdaes,'' \emph{IEEE Transactions on
  Cybernetics (TYCB)}, vol.~46, no.~2, pp. 487--498, 2016.

\bibitem{CPMC2012}
J.~Carreira and C.~Sminchisescu, ``{CPMC: Automatic Object Segmentation Using
  Constrained Parametric Min-Cuts},'' \emph{IEEE Trans. Pattern Anal. Mach.
  Intell. (TPAMI)}, vol.~34, pp. 1312--1328, 2012.

\bibitem{MCG2014}
P.~Arbel\'{a}ez, J.~Pont-Tuset, J.~Barron, F.~Marques, and J.~Malik,
  ``Multiscale combinatorial grouping,'' in \emph{Proc. IEEE Conf. Comput. Vis.
  Pattern Recognit. (CVPR)}, 2014.

\end{thebibliography}


%
\begin{IEEEbiography}[{\includegraphics[width=1in,height=1.25in,clip,keepaspectratio]{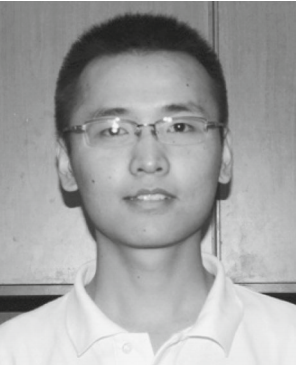}}]{Guanjun Guo}
is currently pursuing the Ph.D. degree with the School of information science technology, Xiamen University, Xiamen, China.
His current research interests include computer vision, machine learning.
\end{IEEEbiography}
\vfill
\begin{IEEEbiography}[{\includegraphics[width=1in,height=1.25in,clip,keepaspectratio]{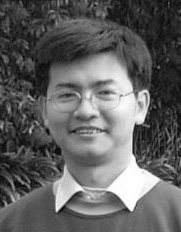}}]{Hanzi Wang}
(SM'10) is currently a Distinguished Professor of ¡±Minjiang Scholars¡± in Fujian province and a Founding Director of the Center for Pattern Analysis and Machine Intelligence (CPAMI) at XMU. He was an Adjunct Professor (2010-2012) and a Senior Research Fellow (2008-2010) at the University of Adelaide, Australia; an Assistant Research Scientist (2007-2008) and a Postdoctoral Fellow (2006-2007) at the Johns Hopkins University; and a Research Fellow at Monash University, Australia (2004-2006). He received his Ph.D degree in Computer Vision from Monash University where he was awarded the Douglas Lampard Electrical Engineering Research Prize and Medal for the best PhD thesis in the Department. His research interests are concentrated on computer vision and pattern recognition including visual tracking, robust statistics, object detection, video segmentation, model fitting, optical flow calculation, 3D structure from motion, image segmentation and related fields. He is a senior member of the IEEE. He was an Associate Editor for IEEE Transactions on Circuits and Systems for Video Technology (from 2010 to 2015). He was the General Chair for ICIMCS2014, Program Chair for CVRS2012, Publicity Chair for IEEE NAS2012, and Area Chair for ACCV2016, DICTA2010. He also serves on the program committee (PC) of ICCV, ECCV, CVPR, ACCV, PAKDD, ICIG, ADMA, CISP, etc, and he serves on the reviewer panel for more than 40 journals and conferences.
\end{IEEEbiography}
\begin{IEEEbiography}[{\includegraphics[width=1in,height=1.25in,clip,keepaspectratio]{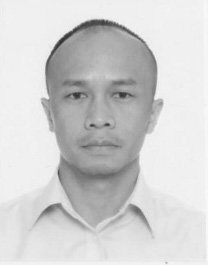}}]{Wan-Lei Zhao}
received the B.Eng. and M.Eng. degrees in computer science and engineering from Yunnan University, Yunnan, China, in 2002 and 2006, respectively, and the Ph.D. degree from the City University
of Hong Kong, Hong Kong, China, in 2010. He is currently with Xiamen University, Xiamen, China, as an Associate Professor. Before joining Xiamen University, he was a Postdoctoral Scholar with
INRIA, Rocquencourt, France. His research interests include multimedia information retrieval and video processing.
\end{IEEEbiography}
\begin{IEEEbiography}[{\includegraphics[width=1in,height=1.25in,clip,keepaspectratio]{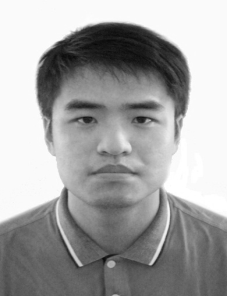}}]{Yan Yan}
(M'13) is currently an associate professor
at the School of Information Science and Technology
at Xiamen University, China. He received the
Ph.D. degree in Information and Communication Engineering
from Tsinghua University, China, in 2009.
He worked at Nokia Japan R\&D center as a research
engineer (2009-2010) and Panasonic Singapore Lab
as a project leader (2011). He has published more
than 30 papers in international journals and conferences
including IEEE T-IP, IEEE T-ITS, PR, KBS,
ICCV, ECCV, ACM MM, etc. His research interests
include computer vision and pattern recognition.
\end{IEEEbiography}
\begin{IEEEbiographynophoto}{Xuelong Li}
(M'02-SM'07-F'12) is a Full Professor with the Center for Optical Imagery Analysis and Learning, State Key Laboratory of Transient
Optics and Photonics, Xi'an Institute of Optics and Precision Mechanics, Chinese Academy of Sciences, Xi'an, China.
\end{IEEEbiographynophoto}
\vfill



\end{document}